%% file: ms.tex
\documentclass[letterpaper, 10 pt, conference]{ieeeconf}  

\input{vcl-shortcuts.tex}

\usepackage{capt-of}
\newcommand{\meas}{\mm}
\newcommand{\act}{\aa}

\newcommand{\obs}{\oo}

\newcommand{\cmd}{\cc}
\newcommand{\cmdset}{\mathcal{C}}

\newcommand{\img}{\ii}
\newcommand{\joint}{\jj}
\newcommand{\latent}{\hh}
\newcommand{\params}{\boldsymbol{\theta}}

\newcommand{\imgmodule}{I}
\newcommand{\measmodule}{M}
\newcommand{\cmdmodule}{C}
\newcommand{\jointmodule}{J}
\newcommand{\actmodule}{A}
\newcommand{\net}{F}
\newcommand{\expert}{E}
\newcommand{\traindata}{\dD}

\newcommand{\loss}{\ell}

\newcommand{\cmdinp}{{\texttt{command input}}}
\newcommand{\branched}{{\texttt{branched}}}
\newcommand{\cmdleft}{{\texttt{left}}}
\newcommand{\cmdstraight}{{\texttt{straight}}}
\newcommand{\cmdright}{{\texttt{right}}}
\newcommand{\cmdcontinue}{{\texttt{continue}}}

\newcommand{\figLabel}{Figure~}
\newcommand{\secLabel}{Section~}

\graphicspath{{./images_compressed/}}

\IEEEoverridecommandlockouts                              

\usepackage{authblk}

\title{\LARGE \bf
End-to-end Driving via Conditional Imitation Learning
}

\author{Felipe Codevilla$^{1,2}$ \and Matthias M\"{u}ller$^{1,3}$ \and Antonio L\'{o}pez$^2$ \and Vladlen Koltun$^1$ \and Alexey Dosovitskiy$^1$}

\begin{document}

\twocolumn[{%
	\renewcommand\twocolumn[1][]{#1}%
	\maketitle
		\vspace{-2mm}
		\centering
 	\begin{tabular}{@{}c@{\hspace{1mm}}c@{\hspace{1mm}}c@{}}
		\includegraphics[height=3.8cm]{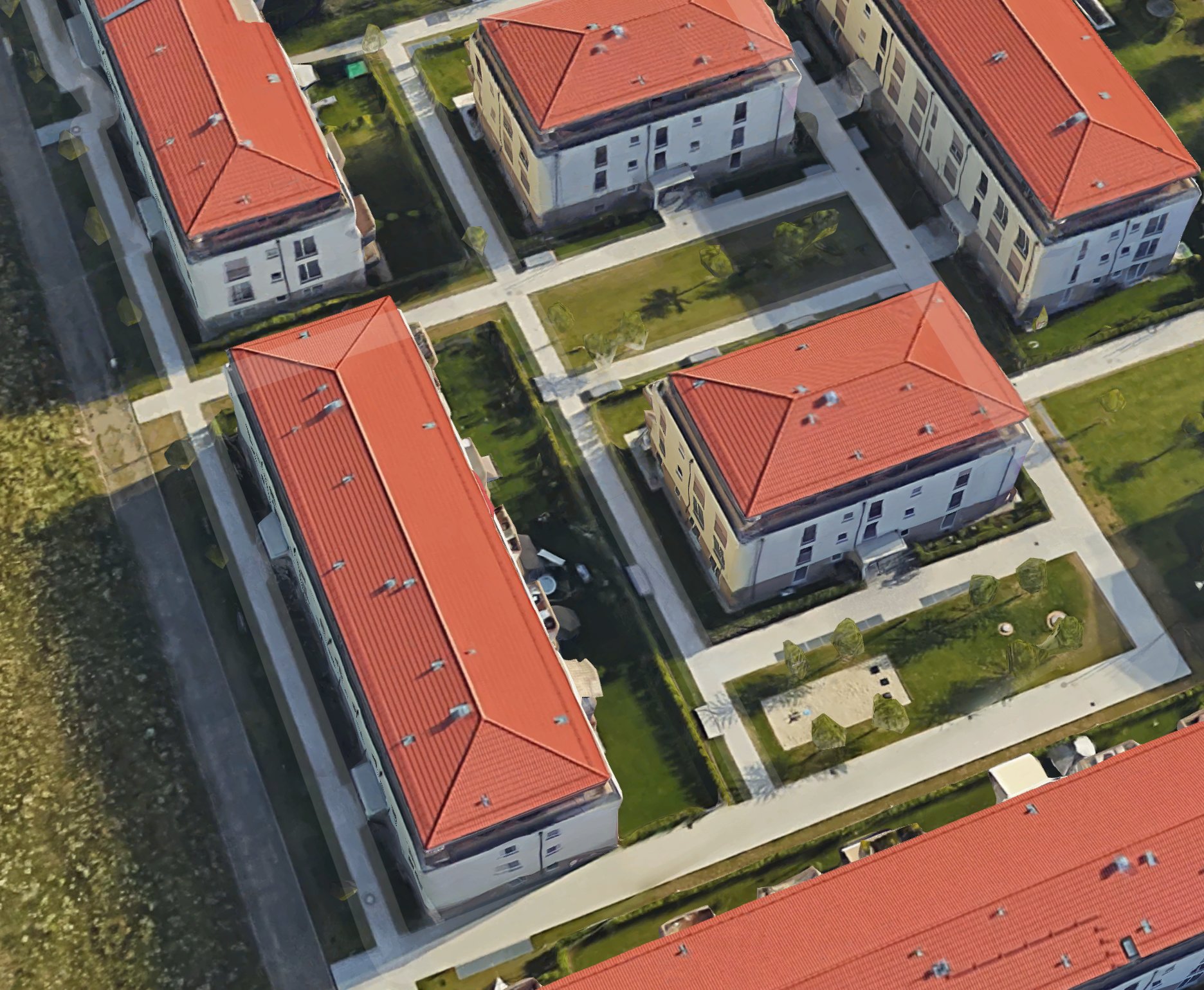} &
		\includegraphics[height=3.8cm]{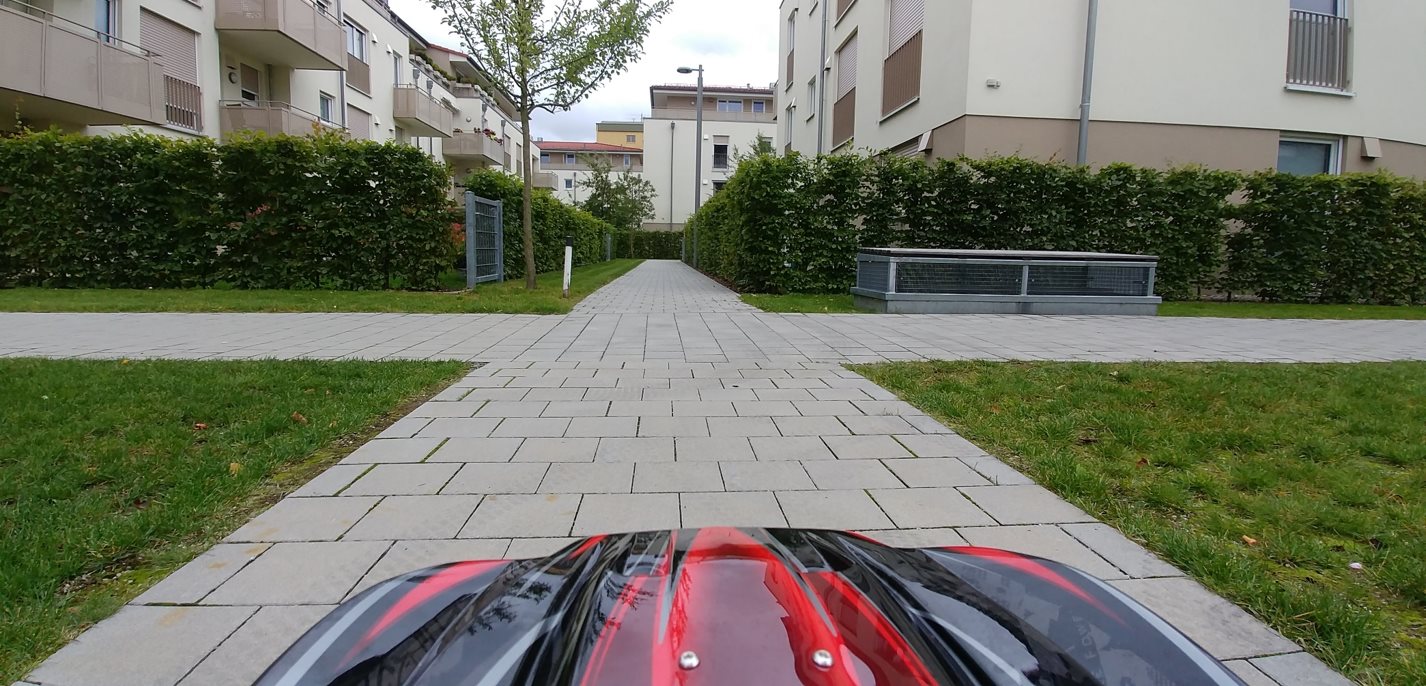} &
		\includegraphics[height=3.8cm]{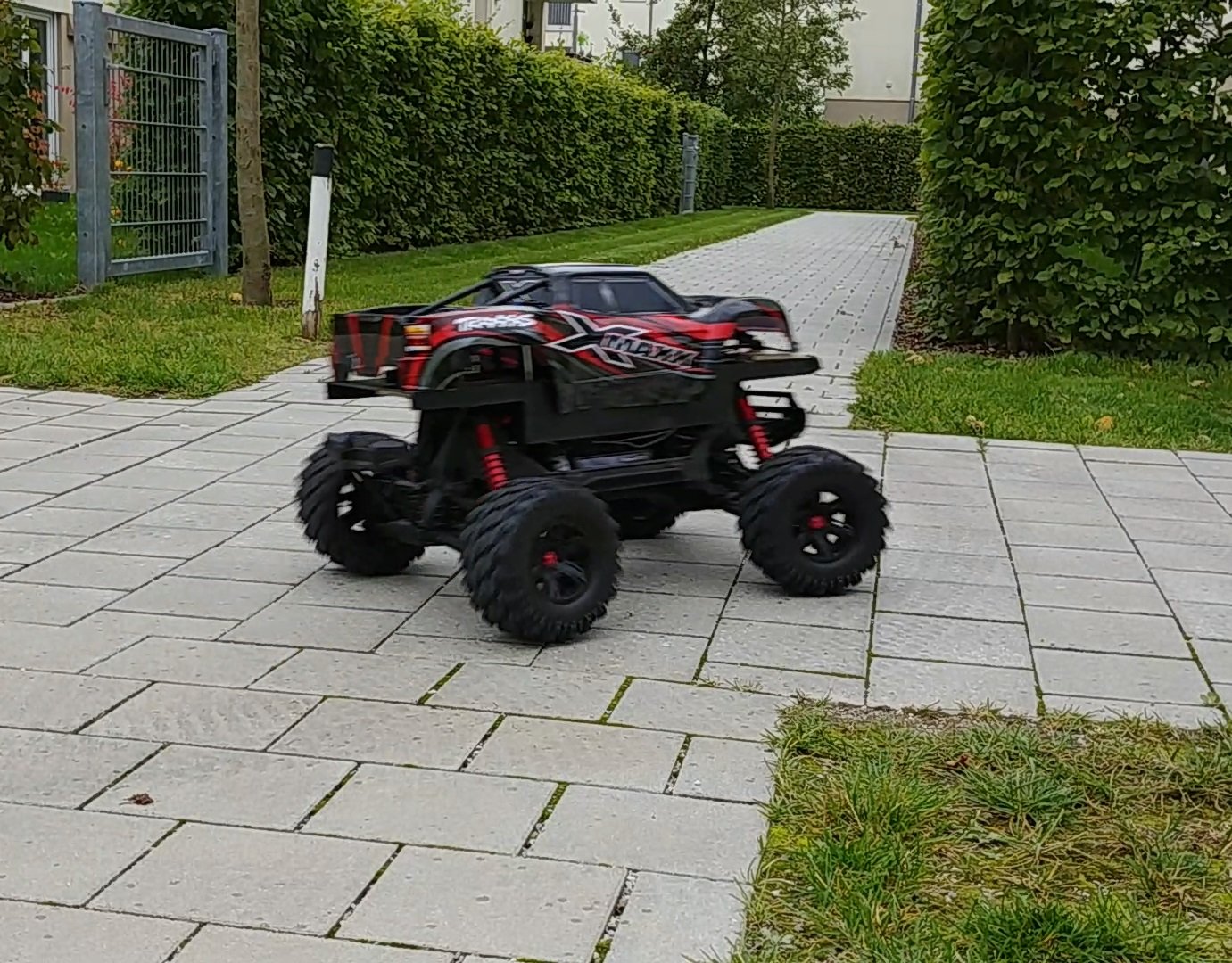}\\
	    \includegraphics[height=3.8cm]{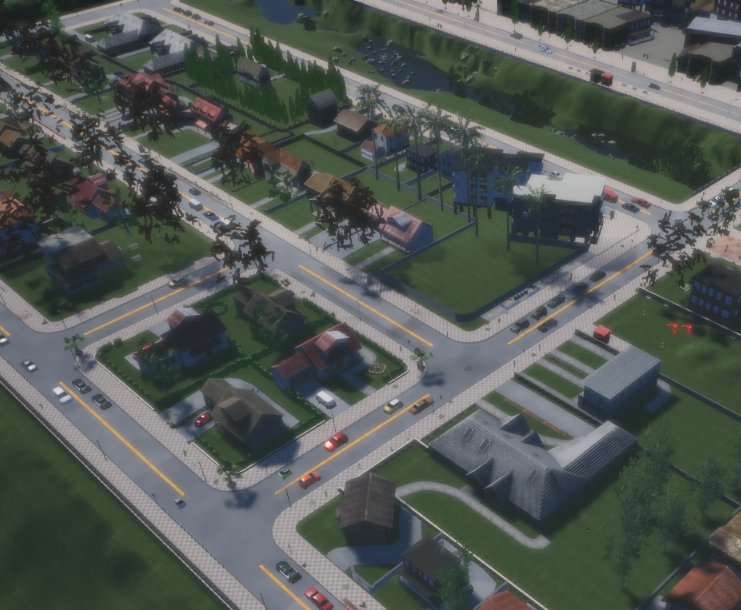} &
		\includegraphics[height=3.8cm]{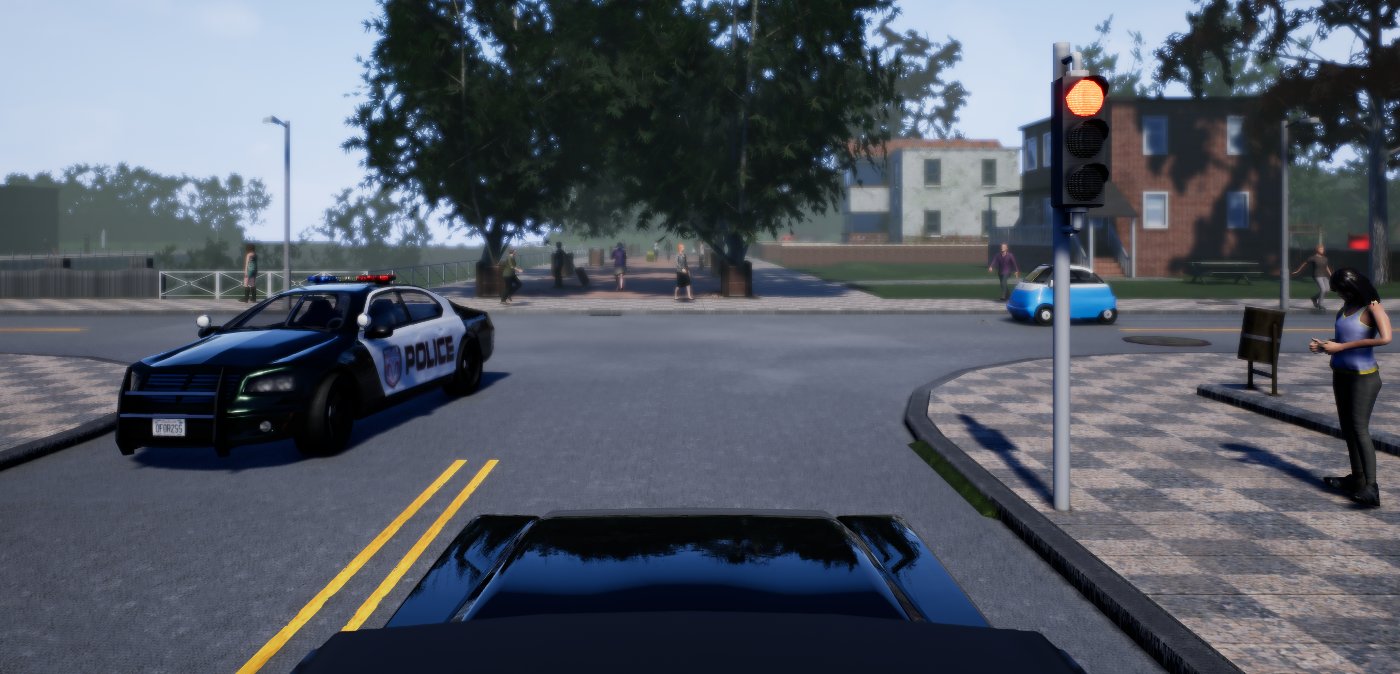} &
		\includegraphics[height=3.8cm]{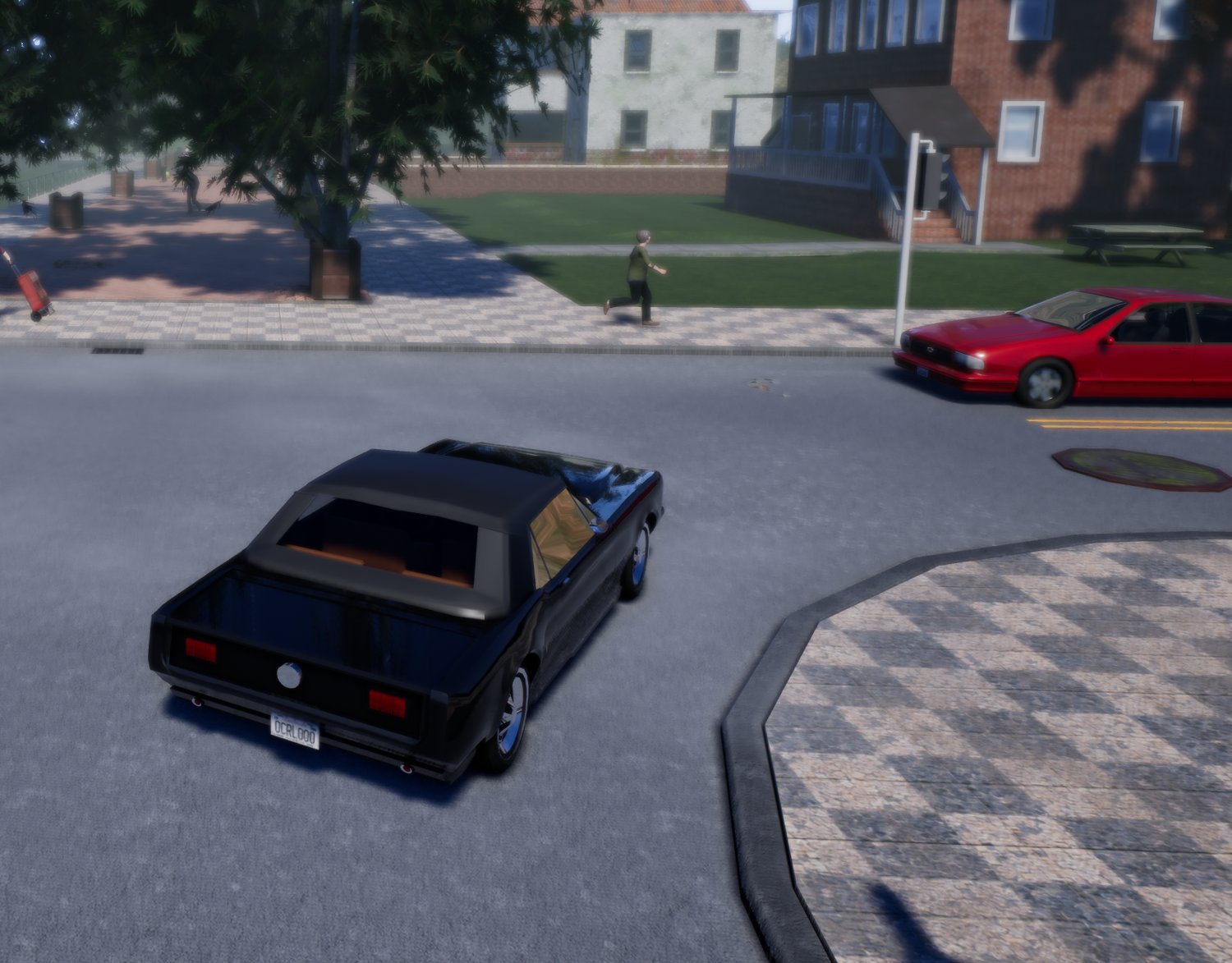}\\
\small (a) Aerial view of test environment & \small (b) Vision-based driving, view from onboard camera & \small (c) Side view of vehicle
\end{tabular}
		\captionof{figure}{Conditional imitation learning allows an autonomous vehicle trained end-to-end to be directed by high-level commands. (a) We train and evaluate robotic vehicles in the physical world (top) and in simulated urban environments (bottom). (b) The vehicles drive based on video from a forward-facing onboard camera. At the time these images were taken, the vehicle was given the command ``turn right at the next intersection''. (c) The trained controller handles sensorimotor coordination (staying on the road, avoiding collisions) and follows the provided commands.}
\label{fig:teaser}
\vspace{3mm}
}]

\thispagestyle{empty}
\pagestyle{empty}

\begin{abstract}
Deep networks trained on demonstrations of human driving have learned to follow roads and avoid obstacles. However, driving policies trained via imitation learning cannot be controlled at test time. A vehicle trained end-to-end to imitate an expert cannot be guided to take a specific turn at an upcoming intersection. This limits the utility of such systems. We propose to condition imitation learning on high-level command input. At test time, the learned driving policy functions as a chauffeur that handles sensorimotor coordination but continues to respond to navigational commands. We evaluate different architectures for conditional imitation learning in vision-based driving. We conduct experiments in realistic three-dimensional simulations of urban driving and on a 1/5 scale robotic truck that is trained to drive in a residential area. Both systems drive based on visual input yet remain responsive to high-level navigational commands.
\end{abstract}

\input{tex/introduction.tex}

\input{tex/background.tex}

\input{tex/commcond.tex}

\input{tex/methodology.tex}

\input{tex/system_setup.tex}

\input{tex/experiments.tex}

\input{tex/conclusions.tex}

\balance

\small
\bibliographystyle{ieee}
\bibliography{biblio}

\end{document}

%% file: vcl-shortcuts.tex
\usepackage{times}
\usepackage{epsfig}
\usepackage{graphicx}
\usepackage{float}
\usepackage{wrapfig}

\usepackage{amsmath,amssymb}

\usepackage{bm,xspace}
\usepackage{comment}
\usepackage{verbatim}
\usepackage{multirow}
\usepackage{balance}
\usepackage{url}
\usepackage{booktabs}
\usepackage{etoolbox,siunitx}
\usepackage{calc}
\usepackage{pifont,hologo}
\usepackage{color}

\newcommand{\ra}[1]{\renewcommand{\arraystretch}{#1}}

\usepackage[super]{nth}
\usepackage{nicefrac}
\def\aa{\mathbf{a}}

\def\cc{\mathbf{c}}

\def\hh{\mathbf{h}}
\def\ii{\mathbf{i}}
\def\jj{\mathbf{j}}

\def\mm{\mathbf{m}}

\def\oo{\mathbf{o}}

\def\tt{\mathbf{t}}

\def\dD{\mathcal{D}}

\DeclareMathOperator*{\minimize}{minimize}

\DeclareMathSymbol{@}{\mathord}{letters}{"3B}

\newcommand\norm[1]{\left\lVert#1\right\rVert}
\newcommand\tuple[1]{\left\langle#1\right\rangle}

\newcommand\timess{\mathbin{\!\times\!}}

\def\latex/{\LaTeX}
\def\bibtex/{\hologo{BibTeX}}


%% file: tex/introduction.tex
\section{Introduction}
\label{sec:introduction}

Imitation learning is receiving renewed interest as a promising approach to training autonomous driving systems.~
\let\thefootnote\relax\footnote{$^1$Intel Labs}
\let\thefootnote\relax\footnote{$^2$Computer Vision Center \& Univ. Autonoma de Barcelona}
\let\thefootnote\relax\footnote{$^3$King Abdullah University of Science \& Technology}
Demonstrations of human driving are easy to collect at scale. Given such demonstrations, imitation learning can be used to train a model that maps perceptual inputs to control commands; for example, mapping camera images to steering and acceleration. This approach has been applied to lane following~\cite{Pomerleau1988,Bojarski2016nvidiadriving} and off-road obstacle avoidance~\cite{LeCun2005driving}. However, these systems have characteristic limitations. For example, the network trained by Bojarski et al.~\cite{Bojarski2016nvidiadriving} was given control over lane and road following only. When a lane change or a turn from one road to another were required, the human driver had to take control~\cite{Bojarski2016nvidiadriving}.

Why has imitation learning not scaled up to fully autonomous urban driving? One limitation is in the assumption that the optimal action can be inferred from the perceptual input alone.
This assumption often does not hold in practice: for instance, when a car approaches an intersection, the camera input is not sufficient to predict whether the car should turn left, right, or go straight.
Mathematically, the mapping from the image to the control command is no longer a function. Fitting a function approximator is thus bound to run into difficulties. This had already been observed in the work of Pomerleau: ``Currently upon reaching a fork, the network may output two widely discrepant travel directions, one for each choice. The result is often an oscillation in the dictated travel direction''~\cite{Pomerleau1988}. Even if the network can resolve the ambiguity in favor of some course of action, it may not be the one desired by the passenger, who lacks a communication channel for controlling the network itself.

In this paper, we address this challenge with conditional imitation learning. At training time, the model is given not only the perceptual input and the control signal, but also a representation of the expert's intention.
At test time, the network can be given corresponding commands, which resolve the ambiguity in the perceptuomotor mapping and allow the trained model to be controlled by a passenger or a topological planner, just as mapping applications and passengers provide turn-by-turn directions to human drivers. The trained network is thus freed from the task of planning and can devote its representational capacity to driving. This enables scaling imitation learning to vision-based driving in complex urban environments.

We evaluate the presented approach in realistic simulations of urban driving and on a 1/5 scale robotic truck. Both systems are shown in Figure~\ref{fig:teaser}.
Simulation allows us to thoroughly analyze the importance of different modeling decisions, carefully compare the approach to relevant baselines, and conduct detailed ablation studies. Experiments with the physical system demonstrate that the approach can be successfully deployed in the physical world. Recordings of both systems are provided in the supplementary video.

%% file: tex/background.tex
\section{Related Work}
\label{sec:background}

Imitation learning has been applied to a variety of tasks, including articulated motion~\cite{Argall2009surveyrobotdemonstration,Ratcliff2007imitationlocomotion,Englert2013}, autonomous flight~\cite{Abbeel2010apprenticeship,Giusti2016,Ross2013uav}, modeling navigational behavior~\cite{Ziebart2008:AAAI,Ziebart2008:Ubicomp}, off-road driving~\cite{LeCun2005driving,Silver2010navigation}, and road following~\cite{Bojarski2016nvidiadriving,Chen:2015,Pomerleau1988,ZhangCho2017}. Technically, these applications differ in the input representation (raw sensory input or hand-crafted features), the control signal being predicted, the learning algorithms, and the learned representations. Most relevant to our work are the systems of Pomerleau~\cite{Pomerleau1988}, LeCun et al.~\cite{LeCun2005driving}, and Bojarski et al.~\cite{Bojarski2016nvidiadriving}, who used ground vehicles and trained deep networks to predict the driver's actions from camera input. These studies focused on purely reactive tasks, such as lane following or obstacle avoidance. In comparison, we develop a command-conditional formulation that enables the application of imitation learning to more complex urban driving. Another difference is that we learn to control not only steering but also acceleration and braking, enabling the model to assume full control of the car.

The decomposition of complex tasks into simpler sub-tasks has been studied from several perspectives. In robotics, movement primitives have been used as building blocks for advanced motor skills~\cite{Kober2013,Pastor2009}. 
Movement primitives represent a simple motion, such as a strike or a throw, by a parameterized dynamical system. In comparison, the policies we consider have much richer parameterizations and address more complex sensorimotor tasks that couple perception and control, such as finding the next opportunity to turn right and then making the turn while avoiding dynamic obstacles.

In reinforcement learning, hierarchical approaches aim to construct multiple levels of temporally extended sub-policies~\cite{BartoMahadevan2003}. The options framework is a prominent example of such hierarchical decomposition~\cite{Sutton1999options}. Basic motor skills that are learned in this framework can be transferred across tasks~\cite{Konidaris2012skilltrees}.
Hierarchical approaches have also been combined with deep learning and applied to raw sensory input~\cite{Kulkarni2016:NIPS}. In these works, the main aim is to learn purely from experience and discover hierarchical structure automatically. This is hard and is in general an open problem, particularly for sensorimotor skills with the complexity we consider. In contrast, we focus on imitation learning, and we provide additional information on the expert's intentions during demonstration. This formulation makes the learning problem more tractable and yields a human-controllable policy.

Adjacent to hierarchical methods is the idea of learning multi-purpose and parameterized controllers. Parameterized goals have been used to train motion controllers in robotics~\cite{daSilva2012,Deisenroth2014,Kober2012}. Schaul et al.~\cite{Schaul2015uvfa} proposed a general framework for reinforcement learning with parameterized value functions, shared across states and goals. Dosovitskiy and Koltun~\cite{Dosovitskiy2017dfp} studied families of parameterized goals in the context of navigation in three-dimensional environments. Javdani et al.~\cite{Javdani2015ahredautonomy} studied a scenario where a robot assists a human and changes its behavior depending on its estimate of the human's goal. Our work shares the idea of training a conditional controller, but differs in the model architecture, the application domain (vision-based autonomous driving), and the learning method (conditional imitation learning).

Autonomous driving is the subject of intensive research~\cite{Paden2016}. Broadly speaking, approaches differ in their level of modularity. On one side are highly tuned systems that deploy an array of computer vision algorithms to create a model of the environment, which is then used for planning and control~\cite{Franke2017}. On the opposite side are end-to-end approaches that train function approximators to map sensory input to control commands~\cite{Bojarski2016nvidiadriving,Pomerleau1988,ZhangCho2017}. Our approach is on the end-to-end side of the spectrum, but in addition to sensory input the controller is provided with commands that specify the driver's intent. This resolves some of the ambiguity in the perceptuomotor mapping and creates a communication channel that can be used to guide the autonomous car as one would guide a chauffeur.

Human guidance of robot actions has been studied extensively~\cite{Broad2017,Hemachandra2015,Matuszek2014,Tellex2011,Walter2013}. These works tackle the challenging problem of parsing natural language instructions. Our work does not address natural language communication; we limit commands to a predefined vocabulary such as ``turn right at the next intersection'', ``turn left at the next intersection'', and ``keep straight''.
On the other hand, our work deals with end-to-end vision-based driving using deep networks. Systems in this domain have been limited to imitating the expert without the ability to naturally accept commands after deployment~\cite{Bojarski2016nvidiadriving,Chen:2015,Pomerleau1988,ZhangCho2017}. We introduce such ability into deep networks for end-to-end vision-based driving.

%% file: tex/commcond.tex
\section{Conditional Imitation Learning}
\label{sec:formulation}

We begin by describing the standard imitation learning setup and then proceed to our command-conditional formulation. Consider a controller that interacts with the environment over discrete time steps. At each time step $t$, the controller receives an observation $\obs_t$ and takes an action $\act_t$.
The basic idea behind imitation learning is to train a controller that mimics an expert. The training data is a set of observation-action pairs $\traindata =\{\tuple{\obs_i, \act_i}\}_{i=1}^N$ generated by the expert. The assumption is that the expert is successful at performing the task of interest and that a controller trained to mimic the expert will also perform the task well. This is a supervised learning problem, in which the parameters $\params$ of a function approximator $\net(\obs; \params)$ must be optimized to fit the mapping of observations to actions:
\begin{equation} \label{eq:imitation}
  \minimize\limits_{\params} \sum_{i} \loss\big(\net(\obs_i; \params), \act_i\big).
\end{equation}

An implicit assumption behind this formulation is that the expert's actions are fully explained by the observations; that is, there exists a function $\expert$ that maps observations to the expert's actions: $\act_i = \expert(\obs_i)$. If this assumption holds, a sufficiently expressive approximator will be able to fit the function $E$ given enough data. This explains the success of imitation learning on tasks such as lane following.
However, in more complex scenarios the assumption that the mapping of observations to actions is a function breaks down. Consider a driver approaching an intersection. The driver's subsequent actions are not explained by the observations, but are additionally affected by the driver's internal state, such as the intended destination. The same observations could lead to different actions, depending on this latent state. This could be modeled as stochasticity, but a stochastic formulation misses the underlying causes of the behavior.
Moreover, even if a controller trained to imitate demonstrations of urban driving did learn to make turns and avoid collisions, it would still not constitute a useful driving system. It would wander the streets, making arbitrary decisions at intersections. A passenger in such a vehicle would not be able to communicate the intended direction of travel to the controller, or give it commands regarding which turns to take.

To address this, we begin by explicitly modeling the expert's internal state by a vector $\latent$, which together with the observation explains the expert's action: $\act_i = \expert(\obs_i, \latent_i)$. Vector $\latent$ can include information about the expert's intentions, goals, and prior knowledge. The standard imitation learning objective can then be rewritten as
\begin{equation} \label{eq:imitation_with_latent}
  \minimize\limits_{\params} \sum_{i} \loss\big(\net(\obs_i; \params), E(\obs_i, \latent_i)\big).
\end{equation}
It is now clear that the expert's action is affected by information that is not provided to the controller $\net$.

We expose the latent state $\latent$ to the controller by introducing an additional command input: $\cmd = \cmd(\latent)$.
At training time, the command $\cmd$ is provided by the expert.
It need not constitute the entire latent state $\latent$, but should provide useful information about the expert's decision-making.
For example, human drivers already use turn signals to communicate their intent when approaching intersections; these turn signals can be used as commands in our formulation.
At test time, commands can be used to affect the behavior of the controller.
These test-time commands can come from a human user or a planning module.
In urban driving, a typical command would be ``turn right at the next intersection'', which can be provided by a navigation system or a passenger.
The training dataset becomes $\traindata = \{\tuple{\obs_i, \cmd_i, \act_i}\}_{i=1}^N$. 

The command-conditional imitation learning objective is
\begin{equation} \label{eq:imitation_with_latent_conditional}
  \minimize\limits_{\params} \sum_{i} \loss\big(\net(\obs_i, \cmd_i; \params), \act_i\big).
\end{equation}

\begin{figure}
	\centering
	\includegraphics[width=1.0\linewidth]{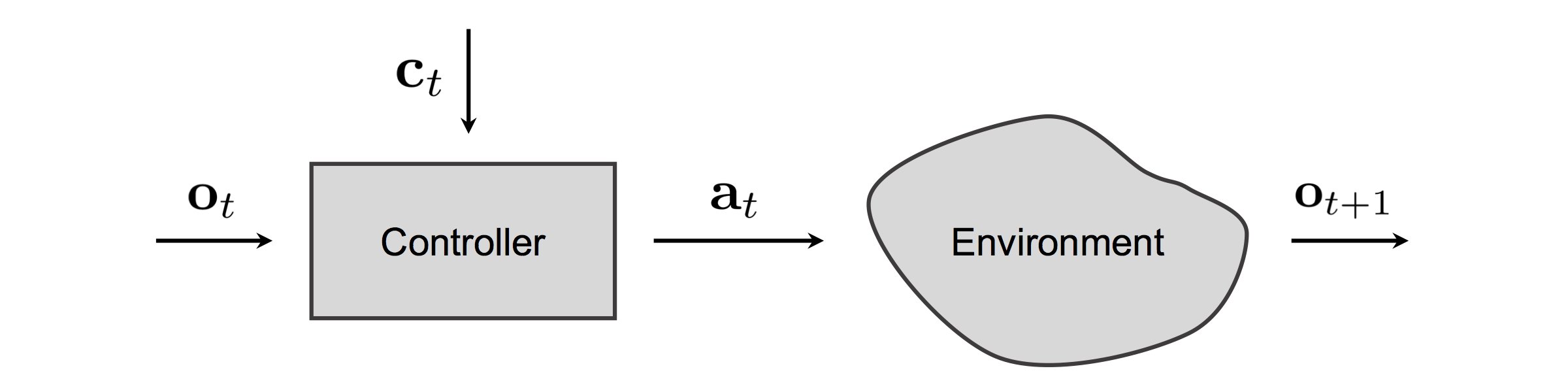}
\caption{High-level overview. The controller receives an observation $\obs_t$ from the environment and a command $\cmd_t$. It produces an action $\act_t$ that affects the environment, advancing to the next time step.}
	\label{fig:high_level}
  \vspace{-3mm}
\end{figure}

In contrast with objective~\eqref{eq:imitation_with_latent}, the learner is informed about the expert's latent state and can use this additional information in predicting the action. This setting is illustrated in Figure~\ref{fig:high_level}.

%% file: tex/methodology.tex
\section{Methodology}

We now describe a practical implementation of command-conditional imitation learning.
Code is available at \url{https://github.com/carla-simulator/imitation-learning}.

\begin{figure*}
	\centering
    \begin{tabular}{c@{\hspace{5mm}}c}
	    \includegraphics[height=3.5cm]{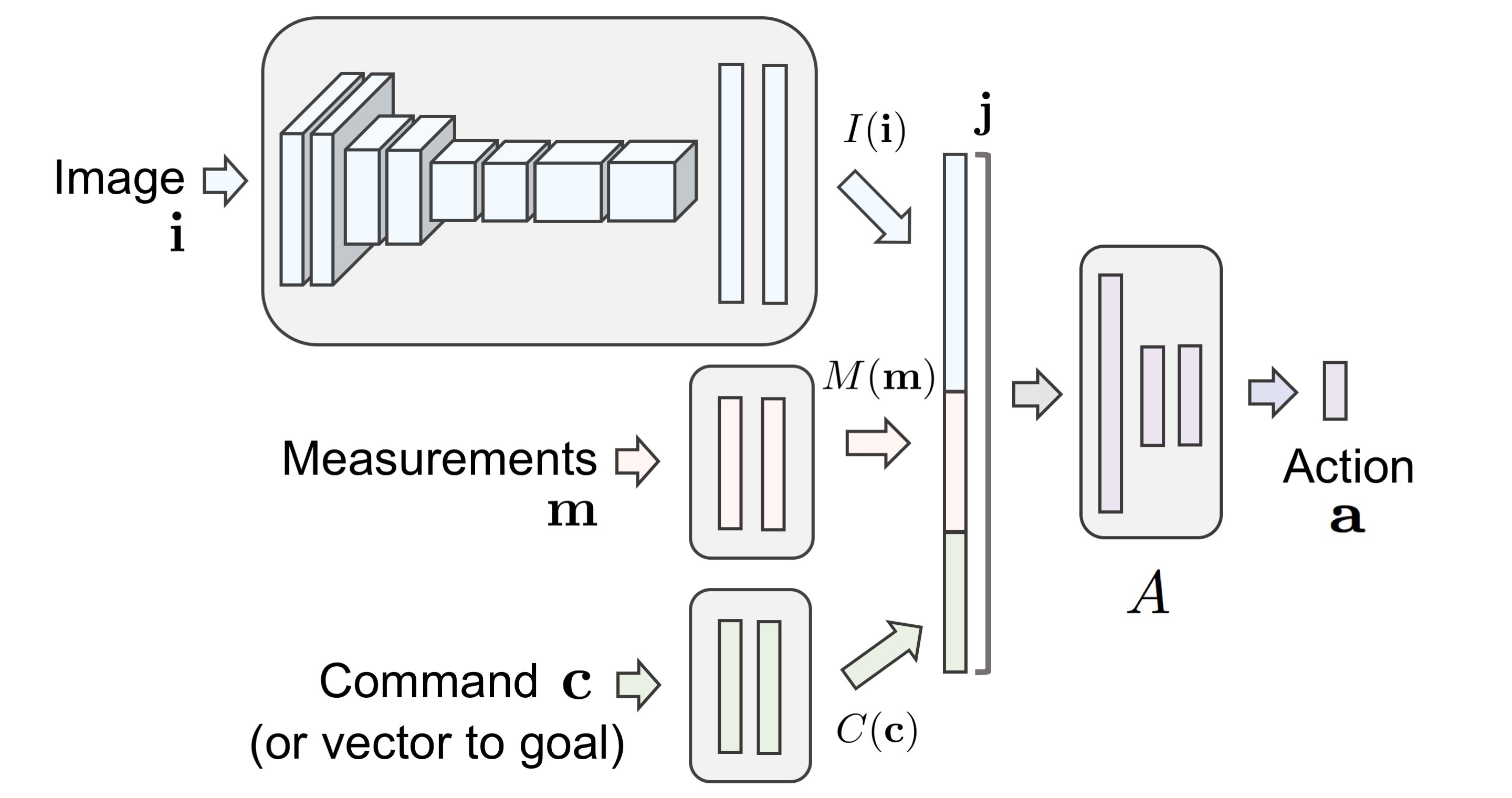} &
      \includegraphics[height=3.5cm]{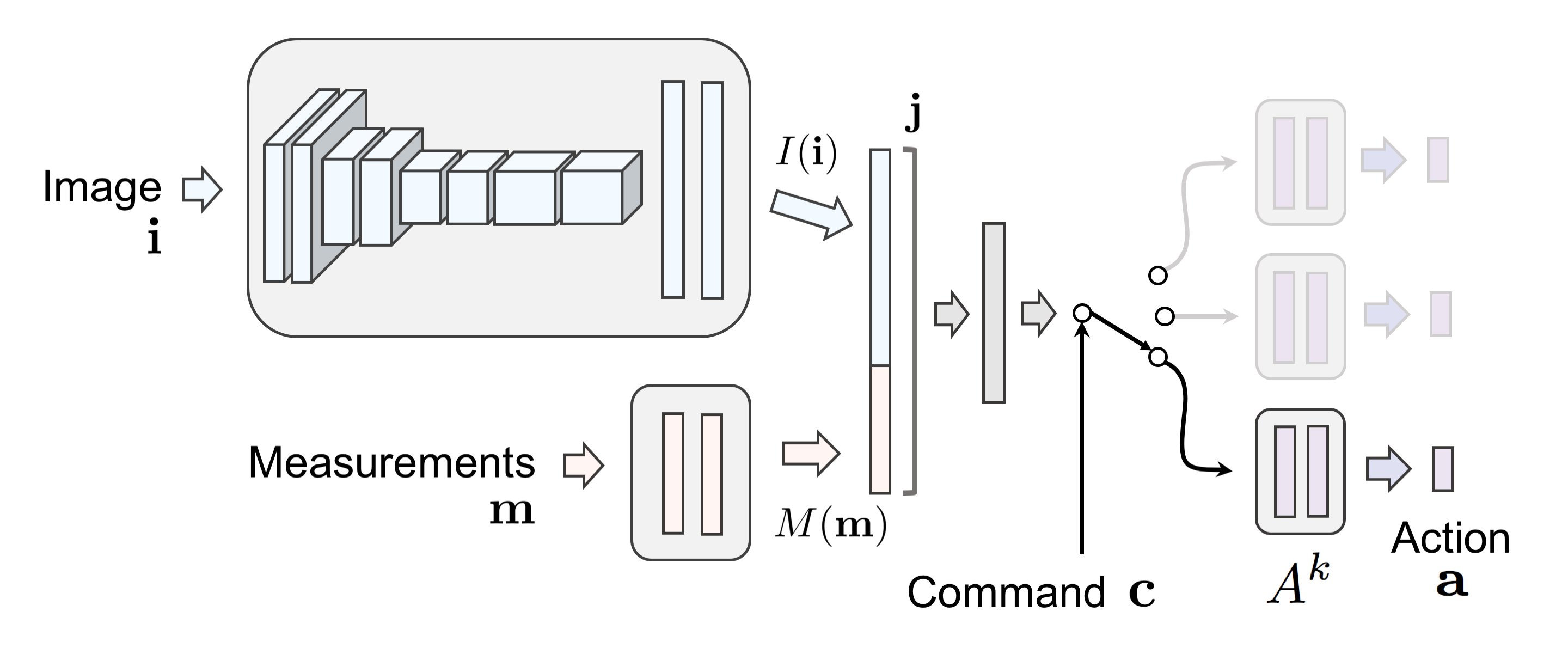} \vspace*{0.1cm} \\
      (a) & (b)
    \end{tabular}
    \caption{Two network architectures for command-conditional imitation learning. (a) \cmdinp{}: the command is processed as input by the network, together with the image and the measurements. The same architecture can be used for goal-conditional learning (one of the baselines in our experiments), by replacing the command by a vector pointing to the goal. (b) \branched{}: the command acts as a switch that selects between specialized sub-modules.}
	\label{fig:net_arch}
  \vspace{-3mm}
\end{figure*}

\subsection{Network Architecture}
\label{sec:architecture}

Assume that each observation $\obs = \tuple{\img, \meas}$ comprises an image $\img$ and a low-dimensional vector $\meas$ that we refer to as measurements, following Dosovitskiy and Koltun~\cite{Dosovitskiy2017dfp}. The controller $\net$ is represented by a deep network. The network takes the image $\img$, the measurements $\meas$, and the command $\cmd$ as inputs, and produces an action $\act$ as its output. The action space can be discrete, continuous, or a hybrid of these. In our driving experiments, the action space is continuous and two-dimensional: steering angle and acceleration.
The acceleration can be negative, which corresponds to braking or driving backwards.
The command $\cmd$ is a categorical variable represented by a one-hot vector.

We study two approaches to incorporating the command $\cmd$ into the network.
The first architecture is illustrated in Figure~\ref{fig:net_arch}(a).
The network takes the command as an input, alongside the image and the measurements.
These three inputs are processed independently by three modules: an image module $\imgmodule(\img)$, a measurement module $\measmodule(\meas)$, and a command module $\cmdmodule(\cmd)$. The image module is implemented as a convolutional network, the other two modules as fully-connected networks. The outputs of these modules are concatenated into a joint representation:
\begin{equation}
  \joint = \jointmodule(\img, \meas, \cmd) = \tuple{\imgmodule(\img), \measmodule(\meas), \cmdmodule(\cmd)} .
\end{equation}
The control module, implemented as a fully-connected network, takes this joint representation and outputs an action $\actmodule(\joint)$.
We refer to this architecture as \cmdinp{}.
It is applicable to both continuous and discrete commands of arbitrary dimensionality.
However, the network is not forced to take the commands into account, which can lead to suboptimal performance in practice.

We therefore designed an alternative architecture, shown in Figure~\ref{fig:net_arch}(b).
The image and measurement modules are as described above, but the command module is removed. Instead, we assume a discrete set of commands ${\cmdset = \{\cmd^0, \ldots, \cmd^K \}}$ (including a default command $\cmd^0$ corresponding to no specific command given) and introduce a specialist branch $\actmodule^i$ for each of the commands $\cmd^i$.
The command $\cmd$ acts as a switch that selects which branch is used at any given time. 
The output of the network is thus
\begin{equation}
  \net(\img, \meas, \cmd^i) = \actmodule^i (\jointmodule(\img, \meas)).
\end{equation}
We refer to this architecture as \branched{}.
The branches $\actmodule^i$ are forced to learn sub-policies that correspond to different commands. In a driving scenario, one module might specialize in lane following, another in right turns, and a third in left turns. All modules share the perception stream.

\subsection{Network Details}
For all controllers, the observation $\oo$ is the currently observed image at $200 \timess 88$ pixel resolution. For the measurement $\mm$, we used the current speed of the car, if available (in the physical system the speed estimates were very noisy and we refrained from using them). All networks are composed of modules with identical architectures (e.g., the ConvNet architecture is the same in all conditions). The differences are in the configuration of modules and branches as can be seen in \figLabel \ref{fig:net_arch}. The image module consists of~8 convolutional and~2 fully connected layers. The convolution kernel size is~5 in the first layer and~3 in the following layers. The first, third, and fifth convolutional layers have a stride of~2. The number of channels increases from $32$ in the first convolutional layer to $256$ in the last. Fully-connected layers contain $512$ units each. All modules with the exception of the image module are implemented as standard multilayer perceptrons. We used ReLU nonlinearities after all hidden layers, performed batch normalization after convolutional layers, applied $50\%$ dropout after fully-connected hidden layers, and used $20\%$ dropout after convolutional layers.

Actions are two-dimensional vectors that collate steering angle and acceleration: ${\act = \tuple{s,a}}$. Given a predicted action $\act$ and a ground truth action $\act_{\text{gt}}$, the per-sample loss function is defined as
\begin{eqnarray}
  \loss (\act, \act_{\text{gt}}) &=& \loss\big(\tuple{s,a},\, \tuple{s_{\text{gt}},a_{\text{gt}}}\big) \nonumber\\
													&=& \norm{s-s_{\text{gt}}}^2 + \lambda_{a} \norm{a - a_{\text{gt}}}^2.
\end{eqnarray}

All models were trained using the Adam solver~\cite{Kingma2015adam} with minibatches of $120$ samples and an initial learning rate of $0.0002$. For the command-conditional models, minibatches were constructed to contain an equal number of samples with each command.

\subsection{Training Data Distribution} \label{sec:method_trainingdata}
When performing imitation learning, a key decision is how to collect the training data. The simplest solution is to collect trajectories from natural demonstrations of an expert performing the task.
This typically leads to unstable policies, since a model that is only trained on expert trajectories may not learn to recover from disturbance or drift~\cite{Levine2013gps,Ross2011dagger}.

To overcome this problem, training data should include observations of recoveries from perturbations.
In DAgger~\cite{Ross2011dagger}, the expert remains in the loop during the training of the controller: the controller is iteratively tested and samples from the obtained trajectories are re-labeled by the expert.
In the system of Bojarski et al.~\cite{Bojarski2016nvidiadriving}, the vehicle is instrumented to record from three cameras simultaneously: one facing forward and the other two shifted to the left and to the right. Recordings from the shifted cameras, as well as intermediate synthetically reprojected views, are added to the training set -- with appropriately adjusted control signals -- to simulate recovery from drift.

\begin{figure}[t]
\vspace{1mm}
\centering
	\begin{tabular}{cc}
	 \begin{tabular}{c}
		 \hspace{-18pt}
	 	\includegraphics[width=0.57\linewidth]{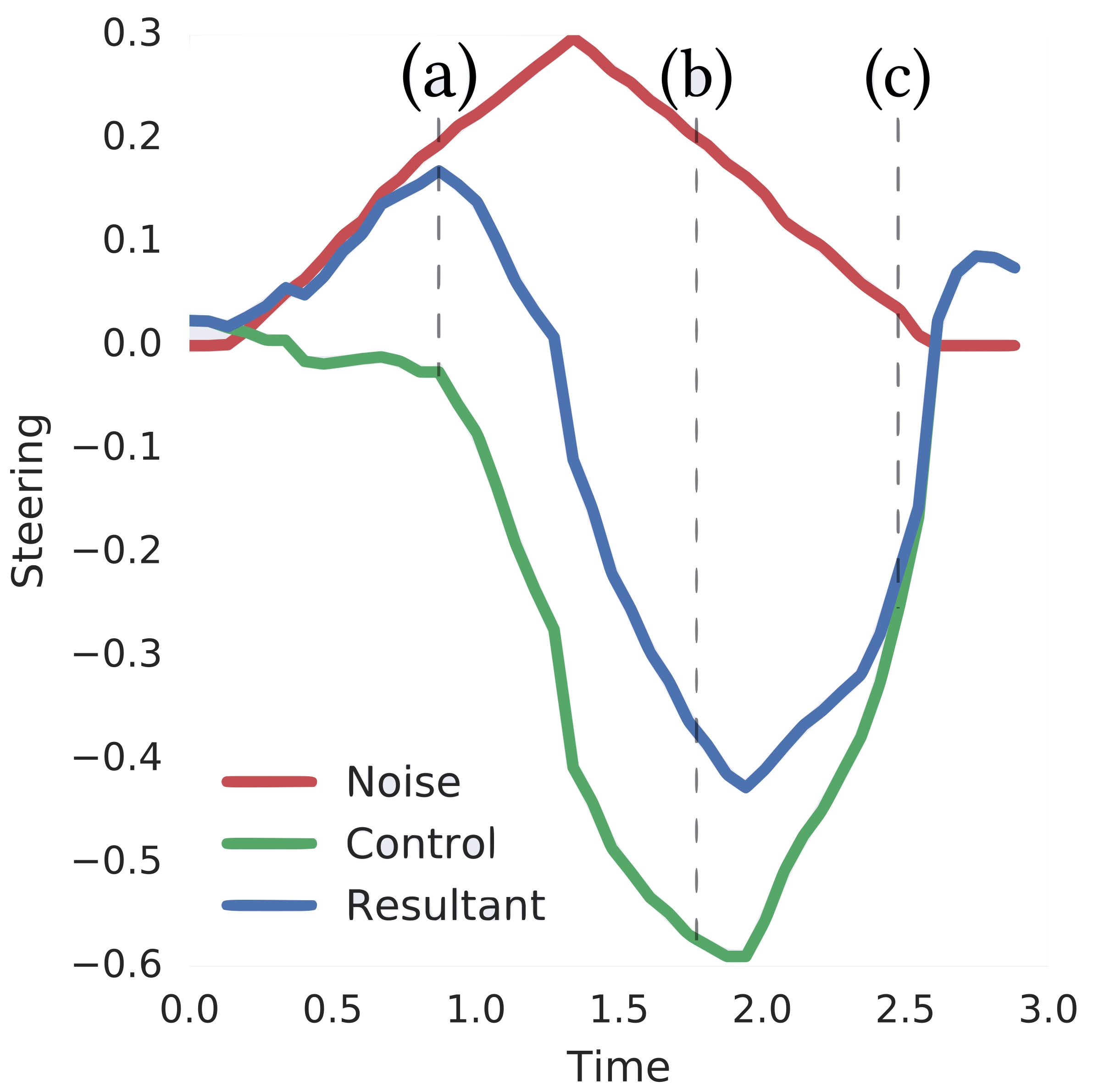}
	 \end{tabular}
	  &
	  \hspace{-0.7cm}
	 \begin{tabular}{c}
	 \includegraphics[width=0.4\linewidth]{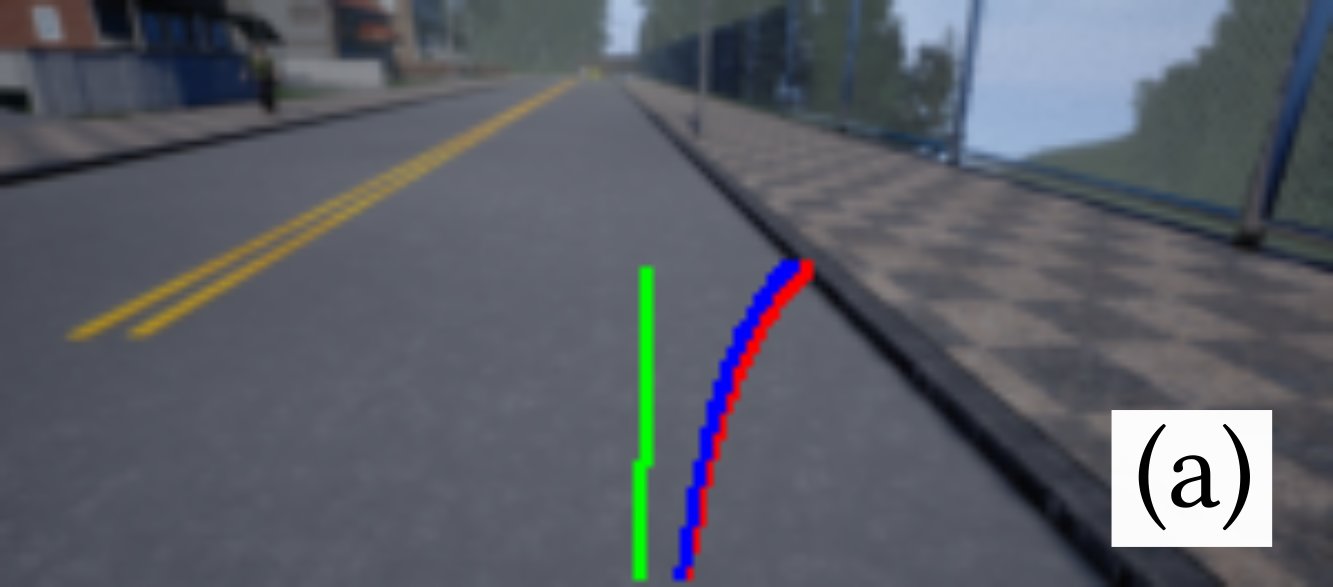} \vspace{-2pt}\\
	 	 \includegraphics[width=0.4\linewidth ]{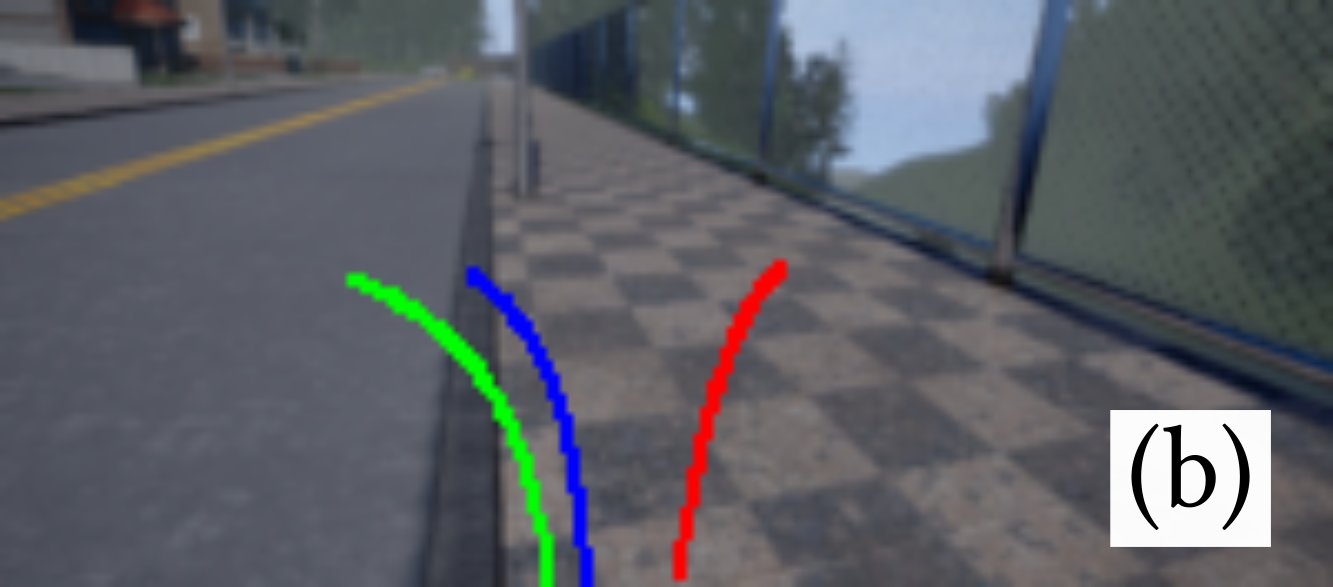}  \vspace{-2pt}\\
	 		 \includegraphics[width=0.4\linewidth ]{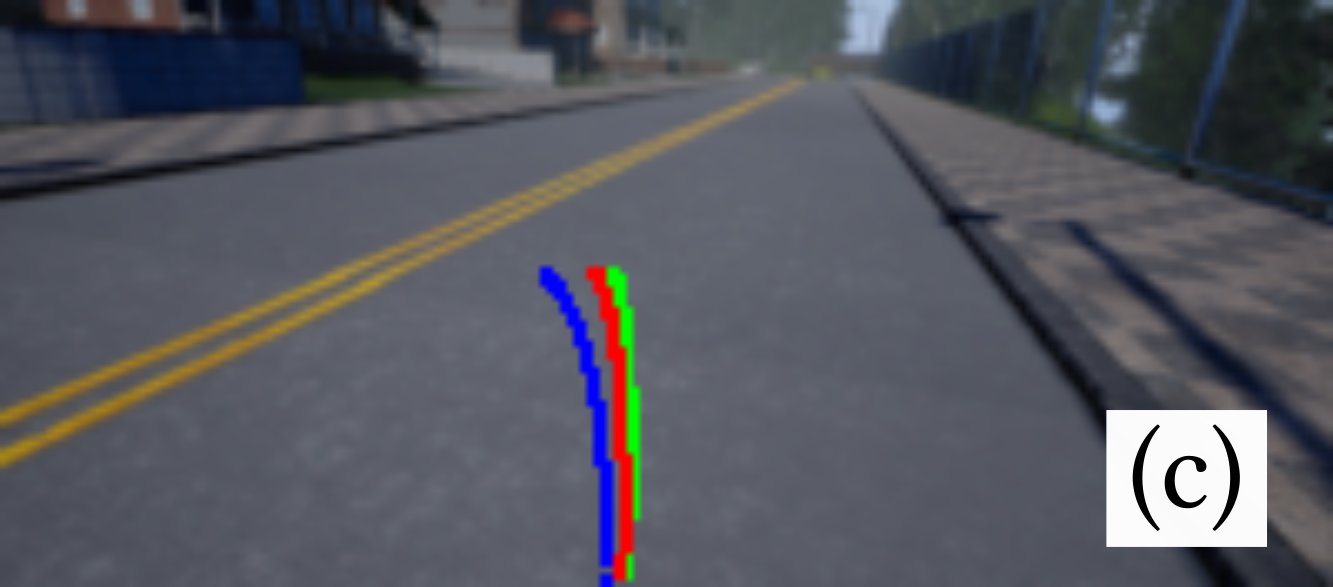} \vspace{-2pt}\\
	 \end{tabular}
	 \end{tabular}
\vspace{-2mm}
    \caption{Noise injection during data collection. We show a fragment from an actual driving sequence from the training set. The plot on the left shows steering control [rad] versus time [s]. In the plot, the red curve is an injected triangular noise signal, the green curve is the driver's steering signal, and the blue curve is the steering signal provided to the car, which is the sum of the driver's control and the noise. Images on the right show the driver's view at three points in time (trajectories overlaid post-hoc for visualization). Between times 0 and roughly 1.0, the noise produces a drift to the right, as illustrated in image (a). This triggers a human reaction, from 1.0 to 2.5 seconds, illustrated in (b). Finally, the car recovers from the disturbance, as shown in (c). Only the driver-provided signal (green curve on the left) is used for training.}
	\label{fig:noise}
	\vspace{-3mm}
\end{figure}

In this paper we adopt a three-camera setup inspired by Bojarski et al.~\cite{Bojarski2016nvidiadriving}.
However, we have found that the policies learned with this setup are not sufficiently robust.
Therefore, to further augment the training dataset, we record some of the data while injecting noise into the expert's control signal and letting the expert recover from these perturbations.
This is akin to the recent approach of Laskey et al.~\cite{Laskey2017noise}, but instead of i.i.d. noise we inject temporally correlated noise designed to simulate gradual drift away from the desired trajectory. An example is shown in Figure~\ref{fig:noise}. For training, we use the driver's corrective response to the injected noise (not the noise itself). This provides the controller with demonstrations of recovery from drift and unexpected disturbances, but does not contaminate the training set with demonstrations of veering away from desired behavior.

\subsection{Data Augmentation}
We found data augmentation to be crucial for good generalization. We perform augmentation online during network training. For each image to be presented to the network, we apply a random subset of a set of transformations with randomly sampled magnitudes. Transformations include change in contrast, brightness, and tone, as well as addition of Gaussian blur, Gaussian noise, salt-and-pepper noise, and region dropout (masking out a random set of rectangles in the image, each rectangle taking roughly 1\% of image area).
No geometric augmentations such as translation or rotation were applied, since control commands are not invariant to these transformations.

%% file: tex/system_setup.tex
\section{System Setup}
\label{sec:system_setup}

We evaluated the presented approach in a simulated urban environment and on a physical system~-- a 1/5 scale truck.
In both cases, the observations (images) are recorded by one central camera and two lateral cameras rotated by 30 degrees
with respect to the center. The recorded control signal is two-dimensional: steering angle and acceleration. The steering angle is scaled between -1 and 1, with extreme values corresponding to full left and full right, respectively.
The acceleration is also scaled between -1 and 1, where 1 corresponds to full forward acceleration and -1 to full reverse acceleration.

In addition to the observations (images) and actions (control signals), we record commands provided by the driver.
We use a set of four commands: \cmdcontinue~(follow the road), \cmdleft~(turn left at the next intersection), \cmdstraight~(go straight at the next intersection), and \cmdright~(turn right at the next intersection). In practice, we represent these as one-hot vectors.

During training data collection, when approaching an intersection the driver uses buttons on a physical steering wheel (when driving in simulation) or on the remote control (when operating the physical truck) to indicate the command corresponding to the intended course of action.
The driver indicates the command when the intended action becomes clear, akin to turn indicators in cars or navigation instructions provided by mapping applications.
This way we collect realistic data that reflects how a higher level planner or a human could direct the system.

\subsection{Simulated Environment}
\begin{figure}[!htb]
	\centering
	\begin{tabular}{@{}c@{\hspace{1mm}}c@{}}
		\includegraphics[width=0.49\columnwidth]{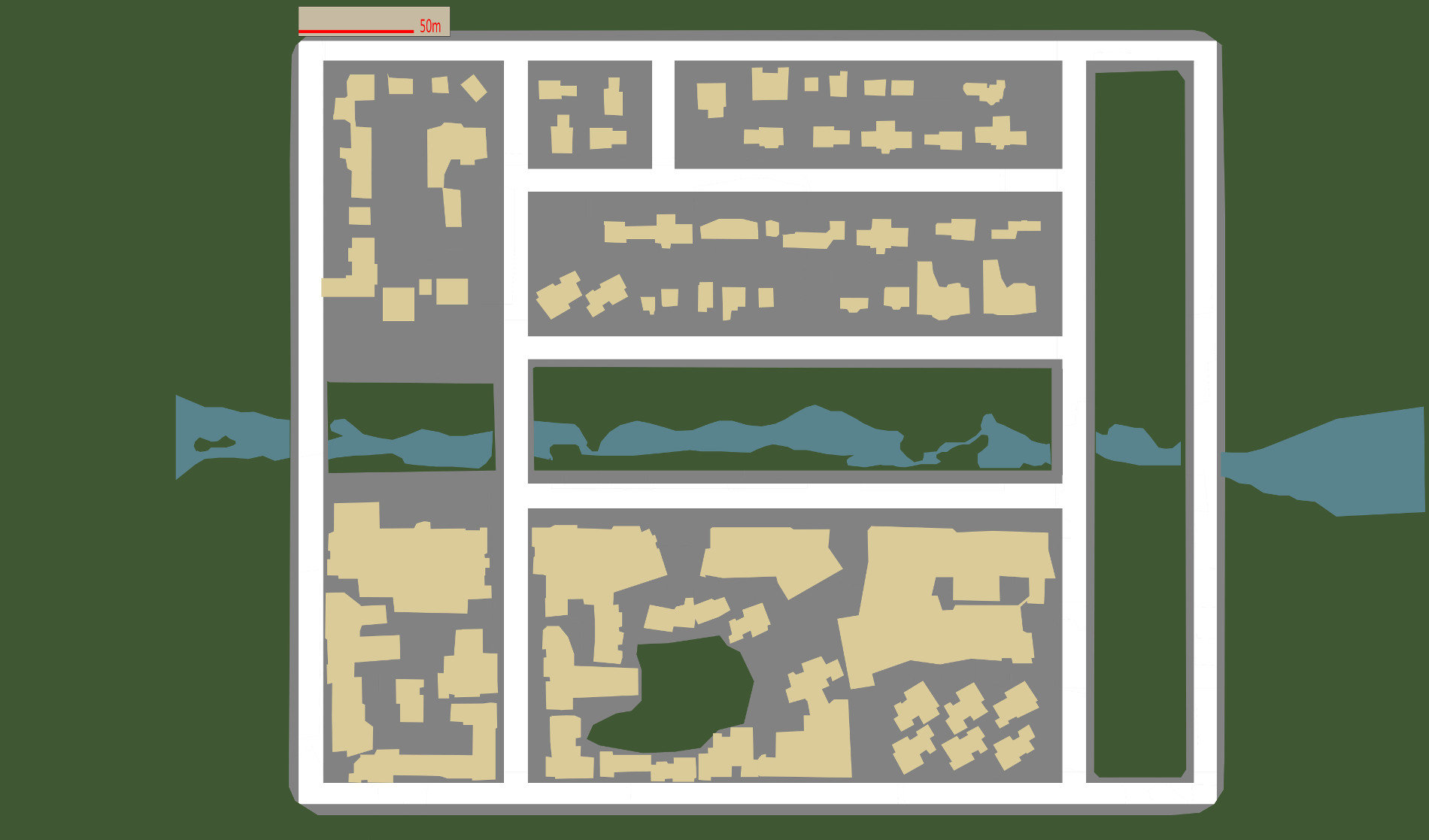} &
		\includegraphics[width=0.49\columnwidth]{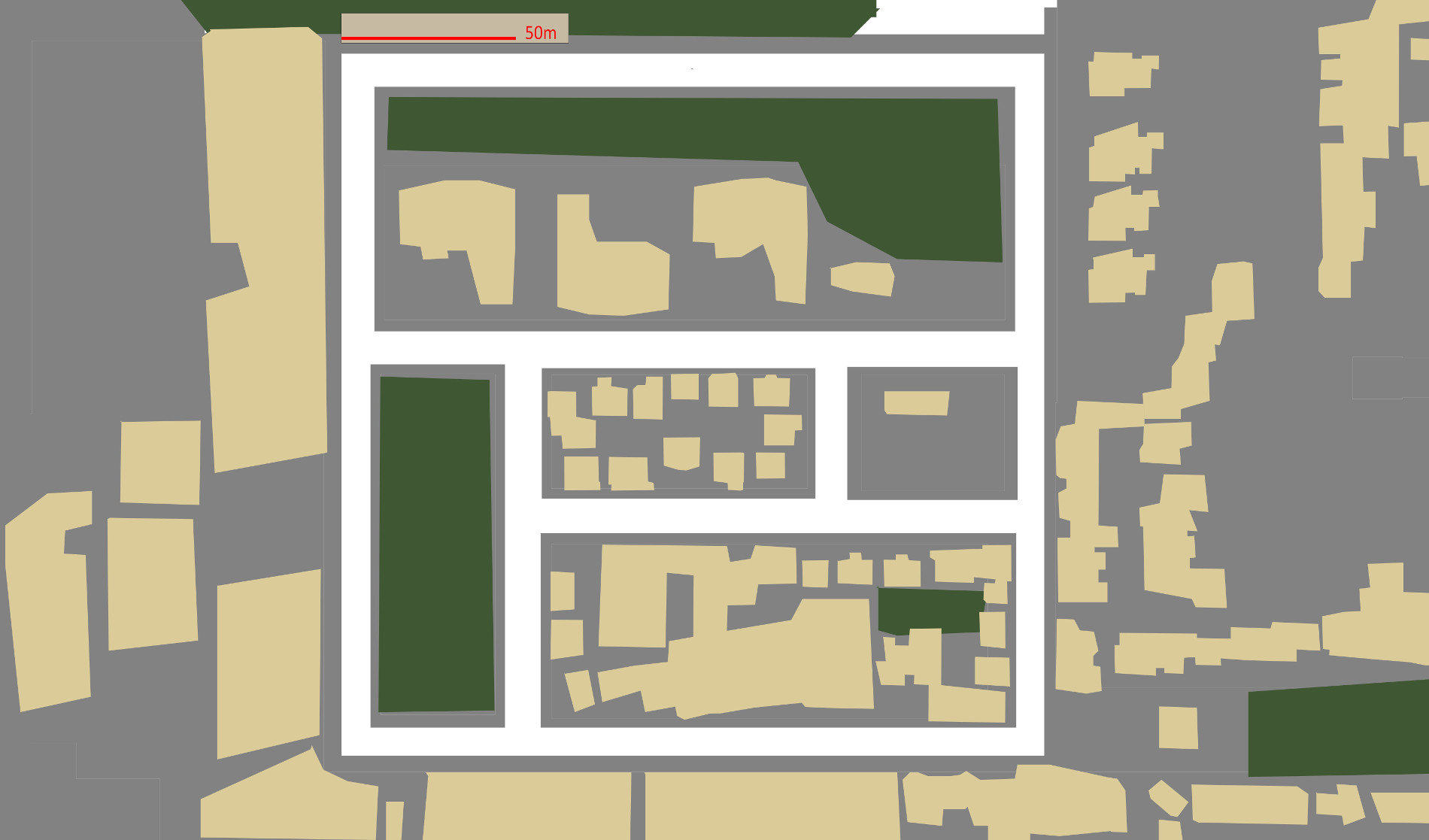} \\
		\includegraphics[width=0.49\columnwidth]{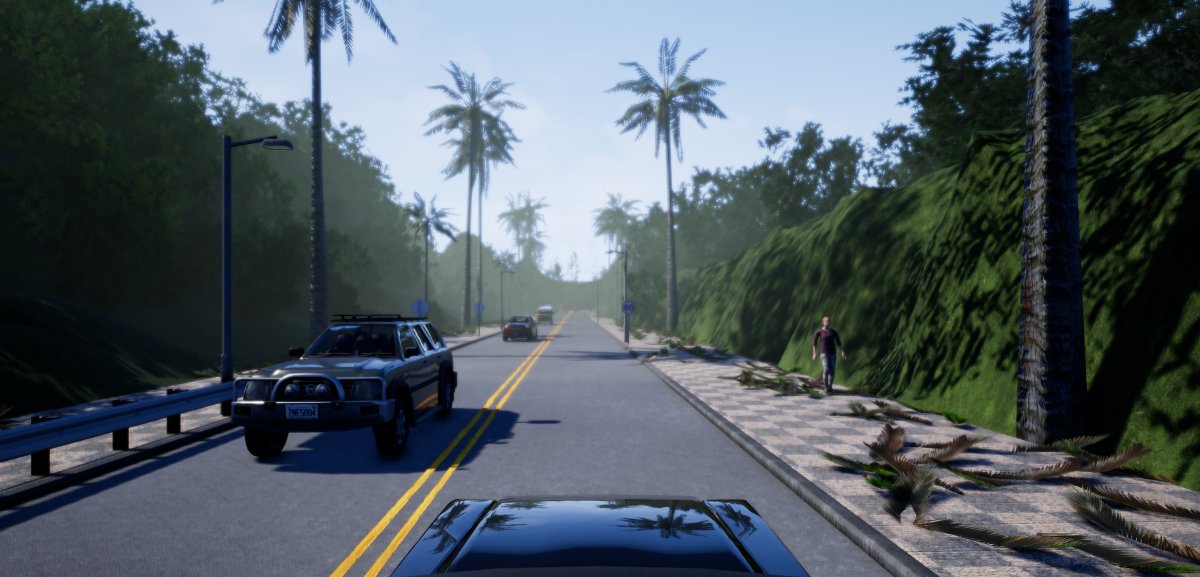} &
		\includegraphics[width=0.49\columnwidth]{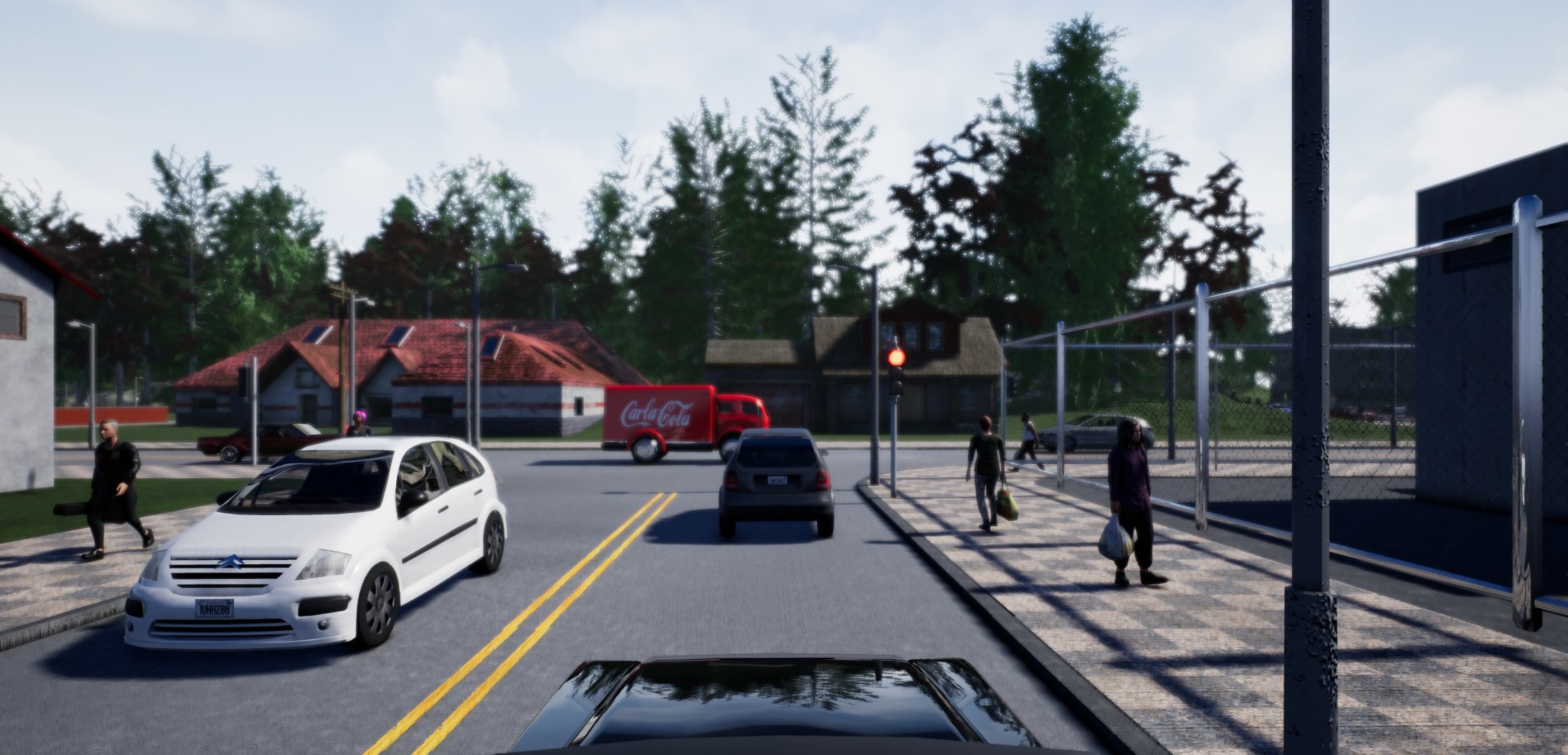}\\
\small Town 1 (training) & \small Town 2 (testing)
	\end{tabular}
	\caption{Simulated urban environments. Town 1 is used for training (left), Town 2 is used exclusively for testing (right). Map on top, view from onboard camera below. Note the difference in visual style.}
	\label{fig:carla}
	\vspace{-1mm}
\end{figure}

We use CARLA~\cite{Dosovitskiy2017}, an urban driving simulator, to corroborate design decisions and evaluate the proposed approach in a dynamic urban environment with traffic.
CARLA is an open-source simulator implemented using Unreal Engine~4. It contains two professionally designed towns with buildings, vegetation, and traffic signs, as well as vehicular and pedestrian traffic. Figure~\ref{fig:carla} provides maps and sample views of Town 1, used for training, and Town 2, used exclusively for testing.

In order to collect training data, a human driver is presented with a first-person view of the environment (center camera) at a resolution of $800 \timess 600$ pixels. The driver controls the simulated vehicle using a physical steering wheel and pedals, and provides command input using buttons on the steering wheel.
The driver keeps the car at a speed below $60$ km/h and strives to avoid collisions with cars and pedestrians, but ignores traffic lights and stop signs.
We record images from the three simulated cameras, along with other measurements such as speed and the position of the car. The images are cropped to remove part of the sky. CARLA also provides extra information such as distance travelled, collisions, and the occurrence of infractions such as drift onto the opposite lane or the sidewalk. This information is used in evaluating different controllers.

\subsection{Physical System}
\label{sec:system_setup_physical}

\begin{figure}[!htb]
	\centering
	\includegraphics[width=\columnwidth]{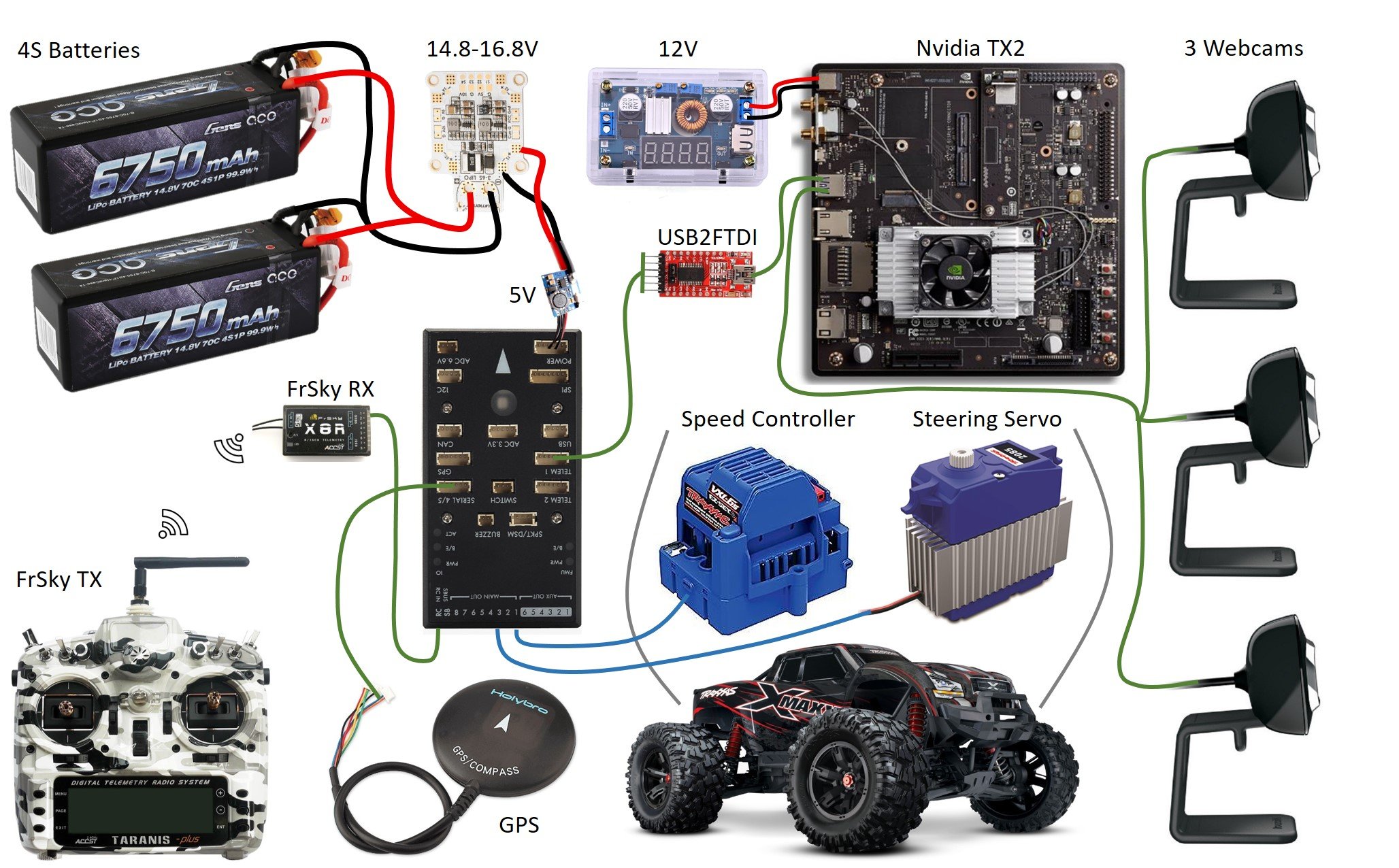}
	\caption{Physical system setup. Red/black indicate +/- power wires, green indicates serial data connections, and blue indicates PWM control signals.}
	\label{fig:physical_system}
	\vspace{-5mm}
\end{figure}

The setup of the physical system is shown in \figLabel \ref{fig:physical_system}. We equipped an off-the-shelf 1/5 scale truck (Traxxas Maxx) with an embedded computer (Nvidia TX2), three low-cost webcams, a flight controller (Holybro Pixhawk) running the APMRover firmware, and supporting electronics. The TX2 acquires images from the webcams and shares a bidirectional communication channel with the Pixhawk. The Pixhawk receives controls from either the TX2 or a human driver and converts them to low-level PWM signals for the speed controller and steering servo of the truck.

\subsubsection{Data collection}
During data collection the truck is driven by a human. The images from all three cameras are synchronized with the control signals and with GPS and IMU data from the Pixhawk, and recorded to disk. The control signals are passed through the TX2 to support noise injection as described in \secLabel \ref{sec:method_trainingdata}. In addition, routing the control through the TX2 ensures a similar delay in the training data as during test time. For the physical system we use only three command inputs (\cmdleft, \cmdstraight, \cmdright), since only a three-way switch is available on the remote control.

\subsubsection{Model evaluation}
\begin{figure}[!htb]
	\centering
	\begin{tabular}{@{}c@{\hspace{1mm}}c@{\hspace{1mm}}c@{}}
		\includegraphics[width=0.32\columnwidth]{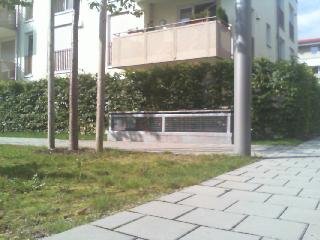} &
    \includegraphics[width=0.32\columnwidth]{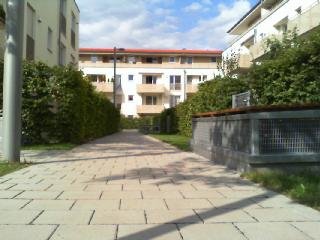} &
    \includegraphics[width=0.32\columnwidth]{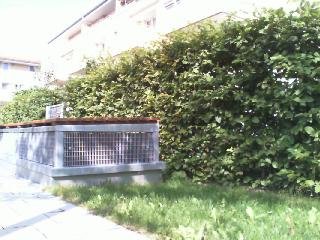}\\
	\small (a) Left camera & \small (b) Central camera & \small (c) Right camera
	\end{tabular}
	\caption{Images from the three cameras on the truck. All three cameras are used for training, with appropriately adjusted steering commands. Only the central camera is used at test time.}
	\label{fig:fpv_car}
	\vspace{-3mm}
\end{figure}

At test time the trained model is evaluated on the TX2 in real time. It receives images from the central webcam and commands (\cmdleft, \cmdstraight, \cmdright) from the remote control. \figLabel \ref{fig:fpv_car}(b) shows an example image from the central camera. The network predicts the appropriate controls in an end-to-end fashion based on only the current image and the provided command. The predicted control is forwarded to the Pixhawk, which controls the car accordingly by sending the appropriate PWM signals to the speed controller and steering servo.

%% file: tex/experiments.tex
\section{Experiments}
\input{tex/experiments_carla.tex}
\input{tex/experiments_rccar.tex}

%% file: tex/experiments_carla.tex
\subsection{Simulated Environment}
\label{sec:experiments_carla}
\subsubsection{Experimental setup}
The use of the CARLA simulator enables running the evaluation in an episodic setup.
In each episode, the agent is initialized at a new location and has to drive to a given destination point, given high-level turn commands from a topological planner.
An episode is considered successful if the agent reaches the goal within a fixed time interval.
In addition to success rate, we measured driving quality by recording the average distance travelled without infractions (collisions or veering outside the lane).

The two CARLA towns used in our experiments are illustrated in Figure~\ref{fig:carla} and in the supplementary video. Town 1 is used for training, Town 2 is used exclusively for testing. For evaluation, we used $50$ pairs of start and goal locations set at least 1 km apart, in each town.

Our training dataset comprises $2$ hours of human driving in Town 1 of which only $10\%$ (roughly $12$ minutes) contain demonstrations with injected noise. Collecting training data with strong injected noise was quite exhausting for the human driver. However, a relatively small amount of such data proved very effective in stabilizing the learned policy.

\begin{table}
\vspace{2mm}
\centering
\ra{1.05}
\small
\resizebox{1\linewidth}{!}{
 \begin{tabular}{@{}lcccccc@{}}
\toprule
             && \multicolumn{2}{c}{Success rate} && \multicolumn{2}{c}{Km per infraction} \\
    Model     && Town 1 & Town 2 && Town 1 & Town 2  \\
    \midrule
    Non-conditional                     && 20\% & 26\%   && 5.76 & 0.89     \\

    Goal-conditional              && 24\% & 30\% && 1.87  & 1.22     \\
\midrule
    \textbf{Ours branched}  && \textbf{88\%} & \textbf{64\%} && 2.34  & 1.18     \\
\midrule
    Ours cmd. input     && 78\% & 52\% && 3.97  & 1.30    \\
    Ours no noise                   && 56\% & 22\% && 1.31  & 0.54   \\
    Ours no aug.                   && 80\% & 0\% && 4.03  & 0.36     \\
    Ours shallow net                  && 46\% & 14\% && 0.96  & 0.42   \\
\bottomrule
 \end{tabular}
}

\caption{Results in the simulated urban environment. We compare the presented method to baseline approaches and perform an ablation study.
We measure the percentage of successful episodes and the average distance (in km) driven between infractions.
Higher is better in both cases, but we rank methods based on success.
The proposed \branched{} architecture outperforms the baselines and the ablated versions.
}
\label{tbl:baselines}
\vspace{-5mm}
\end{table}

\subsubsection{Results}
We compare the \branched{} command-conditional architecture, as shown in Figure~\ref{fig:net_arch}(b), with two baseline approaches, as well as several ablated versions of the full architecture.
The two baselines are standard imitation learning and  goal-conditioned imitation learning.
In standard (non-conditional) imitation learning, the action $\aa$ is predicted from the observation $\oo$ and the measurement $\mm$.
In the goal-conditional variant, the controller is additionally provided with a vector pointing to the goal, in the car's coordinate system (the architecture follows Figure~\ref{fig:net_arch}(a)).
Ablated versions include: a network with the \cmdinp{} architecture instead of \branched{} (see Figure~\ref{fig:net_arch}), and three variants of the \branched{} network: trained without noise-injected data, trained without data augmentation, and implemented with a shallower network.

The results are summarized in Table~\ref{tbl:baselines}.
The controller that is trained using standard imitation learning only completes $20\%$ of the episodes in Town 1 and $24\%$ in Town 2, which is not surprising given its ignorance of the goal.
More interestingly, the goal-conditional controller, which is provided with an accurate vector to the goal at every time step during both training and at test time, is performing only slightly better than the non-conditional controller, successfully completing $24\%$ of the episodes in Town 1 and $30\%$ in Town 2. Qualitatively, this controller eventually veers off the road attempting to shortcut to the goal. This also decreases the number of kilometers the controller is able to traverse without infractions. A simple feed-forward network does not automatically learn to convert a vector pointing to the goal into a sequence of turns.

The proposed \branched{} command-conditional controller performs significantly better than the baseline methods in both towns, successfully completing $88\%$ of the episodes in Town 1 and $64\%$ in Town 2.
In terms of distance travelled without infractions, in Town 2 the method is on par with baselines, while in Town 1 it is outperformed by the non-conditional model. This difference is misleading: the non-conditional model drives more cleanly because it is not constrained to travel towards the goal and therefore typically takes a simpler route at each intersection.

The ablation study shown in the bottom part of Table~\ref{tbl:baselines} reveals that all components of the proposed system are important for good performance.
The \branched{} architecture reaches the goal more reliably than the \cmdinp{} one.
The addition of even a small amount of training data with noise in the steering dramatically improves the performance. (Recall that we have only $12$ minutes of noisy data out of the total of $2$ hours.)
Careful data augmentation is crucial for generalization, even within Town 1, but much more so in the previously unseen Town 2: the model without data augmentation was not able to complete a single episode there.
Finally, a sufficiently deep network is needed to learn the perceptuomotor mapping in the visually rich and complex simulated urban environment.

%% file: tex/experiments_rccar.tex
\subsection{Physical System}

\subsubsection{Experimental setup}
\begin{figure}
\vspace{2mm}
	\centering
	\includegraphics[width=0.75\columnwidth]{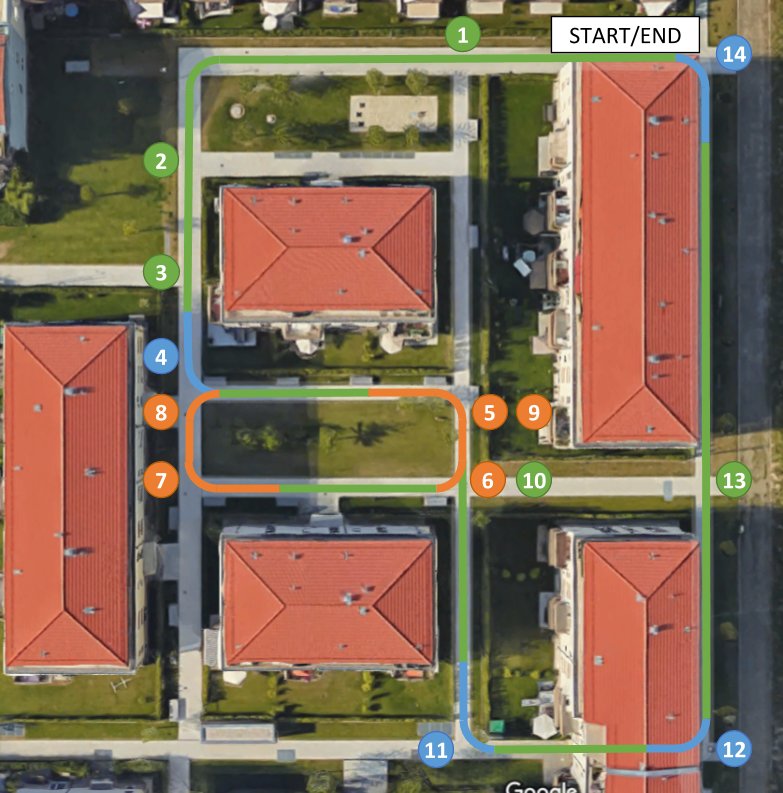}
	\caption{A map of the primary route used for testing the physical system. Intersections traversed by the truck are numbered according to their order along the route. Colors indicate commands provided to the vehicle when it approaches the intersection: blue = \cmdleft, green = \cmdstraight, orange = \cmdright.}
	\label{fig:physical_system_map2D}
	\vspace{-6mm}
\end{figure}

The training dataset consists of $2$ hours of driving the truck via remote control in a residential area.
\figLabel \ref{fig:physical_system_map2D} shows a map with the route on which the vehicle was evaluated. The route includes a total of 14 intersections with roughly the same number of \cmdleft, \cmdstraight, and \cmdright.

We measure the performance in terms of missed intersections, interventions, and time to complete the course. If the robotic vehicle misses an intersection for the first time, it is rerouted to get a second chance to do the turn. If it manages to do the turn the second time, this is not counted as a missed intersection but increases the time taken to complete the route. However, if the vehicle misses the intersection for the second time, this is counted as missed and we intervene to drive the vehicle through the turn manually. Besides missed intersections, we also intervene if the vehicle goes off the road for more than five seconds or if it collides with an obstacle. The models were all evaluated in overcast weather conditions. The majority of training data was collected in sunny weather.

\subsubsection{Main results}
\label{sec:physical_main_results}
We select the most important comparisons from the extensive evaluation performed in simulation (Section~\ref{sec:experiments_carla}) and perform them on the physical system. Table \ref{tbl:physical_system_ablation} shows the results of several variants of command-conditional imitation learning: \branched{} and \cmdinp{} architectures, as well as two ablated models, trained without data augmentation or without noise-injected data. It is evident that the \branched{} architecture achieves the best performance. The ablation experiments show the impact of our noise injection method and augmentation strategy. The model trained without noise injection is very unstable, as indicated by the average number of interventions rising from $0.67$ to $8.67$. Moreover, it misses almost $25\%$ of the intersections and takes double the time to complete the course. The model trained without data augmentation fails completely. The truck misses most intersections and very frequently leaves the lane resulting in almost $40$ interventions. It takes more than four times longer to complete the course. This extreme degradation highlights the importance of generalization in real world settings with constantly changing environmental conditions such as weather and lighting. Proper data augmentation dramatically improves performance given limited training data.

\begin{table}[h]
	\centering
	\ra{1.05}
{\small
		\begin{tabular}{@{}l@{\hspace{4mm}}c@{\hspace{4mm}}c@{\hspace{4mm}}r@{}}
		\toprule
		Model     & Missed turns & Interventions & Time \\
		\midrule
		\textbf{Ours branched}          & \textbf{0\%}          & \textbf{0.67}       & \textbf{2:19}   \\
		Ours cmd. input   & 11.1\%       & 2.33       & 4:13   \\
		Ours no noise           & 24.4\%          & 8.67       & 4:39   \\
		Ours no aug.    & 73\%         & 39         & 10:41  \\
		\bottomrule
		\end{tabular}
	}
	\captionof{table}{Results on the physical system. Lower is better. We compare the \branched{} model to the simpler \cmdinp{} architecture and to ablated versions (without noise injection and without data augmentation). Average performance across 3 runs is reported for all models except for ``Ours no aug.'', for which we only performed 1 run to avoid breaking the truck.}
	\label{tbl:physical_system_ablation}
	\vspace{-1mm}
\end{table}

\subsubsection{Generalization to new environments}
\begin{figure}[h]
\vspace{2mm}
	\centering
	\begin{tabular}{@{}c@{\hspace{1mm}}c@{\hspace{1mm}}c@{}}
		\includegraphics[width=0.32\columnwidth]{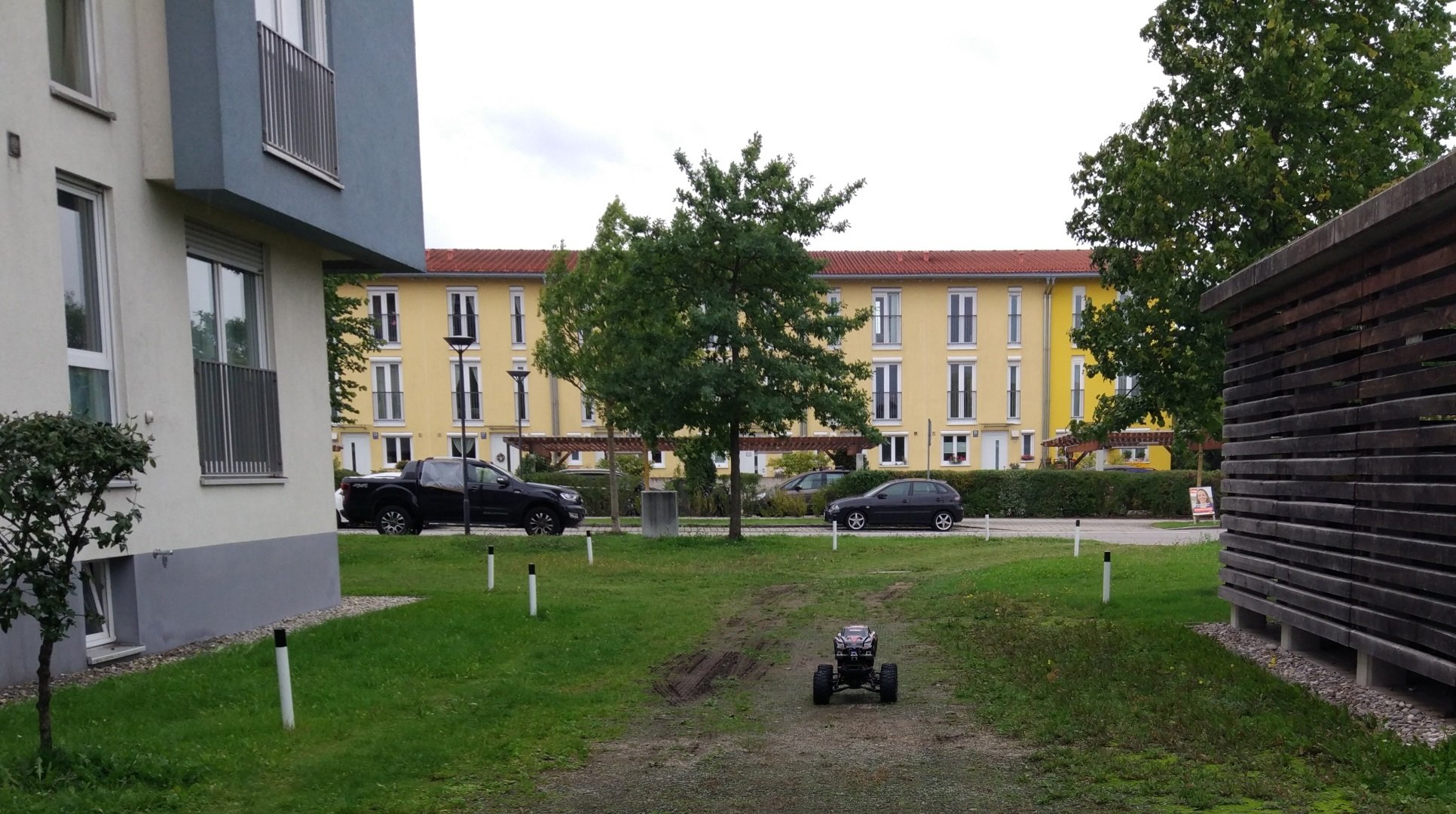} &
		\includegraphics[width=0.32\columnwidth]{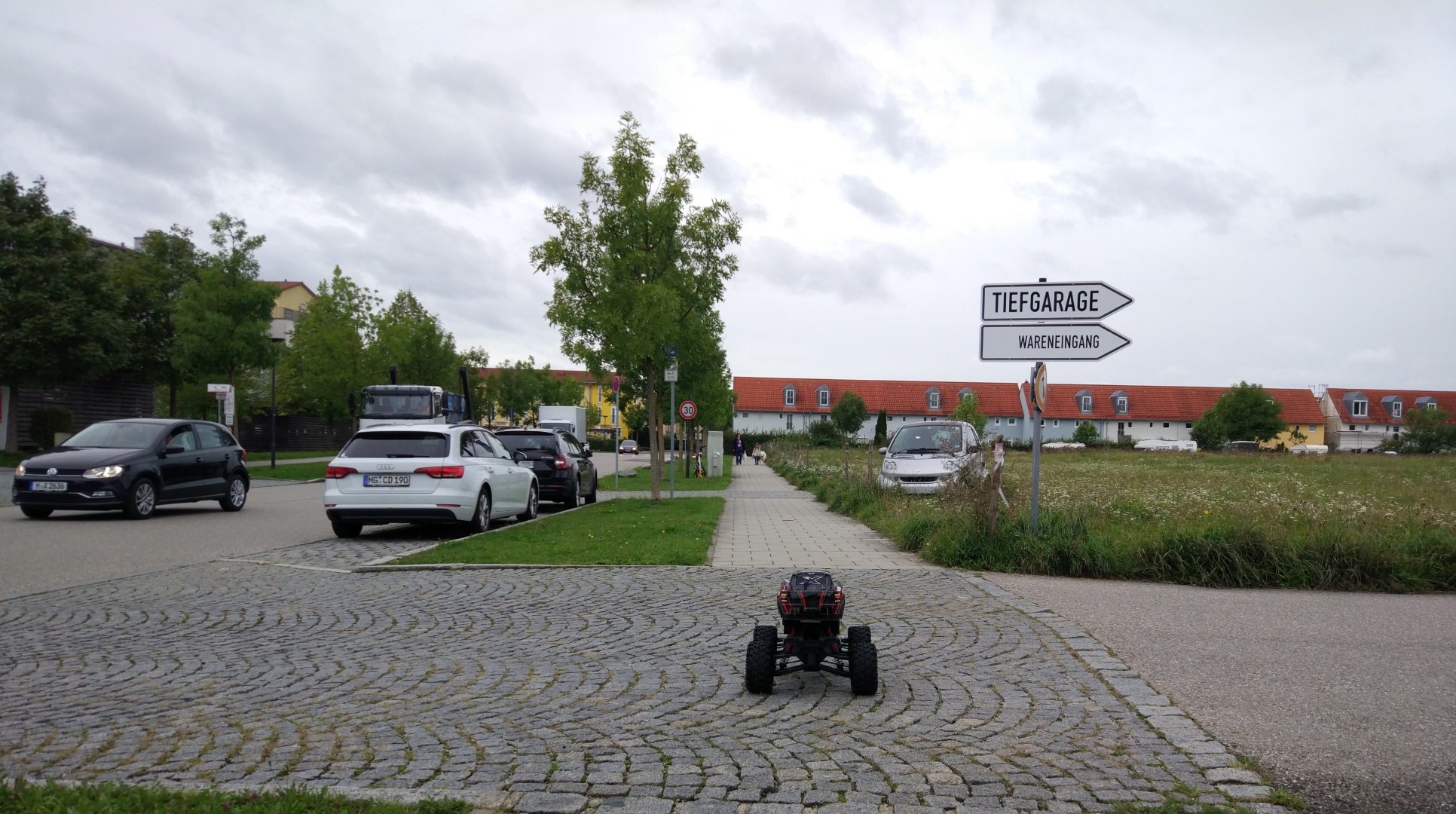} &
		\includegraphics[width=0.32\columnwidth]{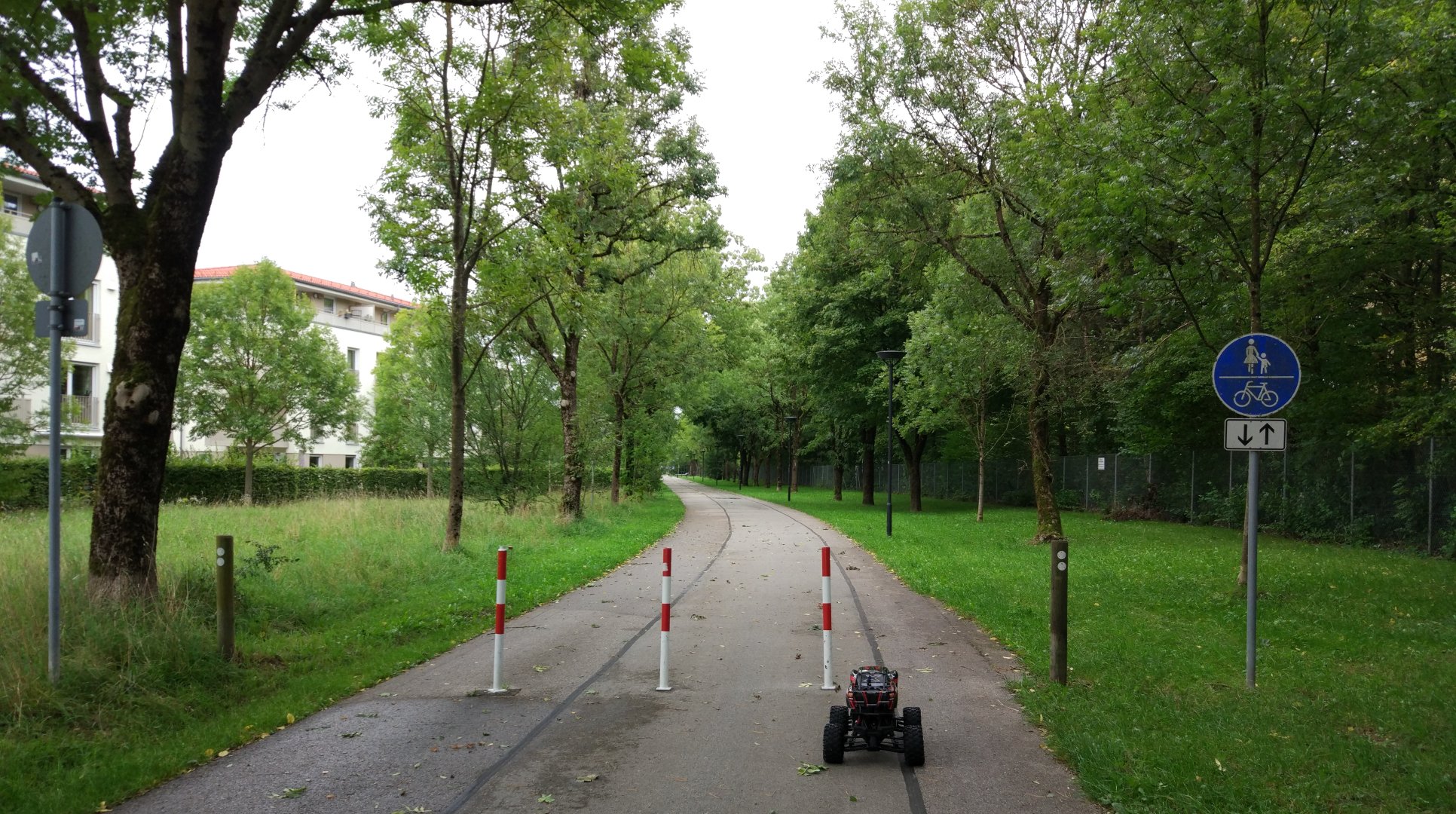}
	\end{tabular}
	\caption{Testing in new environments with very different appearance.}
	\label{fig:physical_generalization}
	\vspace{-3mm}
\end{figure}

Beyond the implicit generalization to varying weather conditions that occur naturally in the physical world, we also evaluate qualitatively how well the model generalizes to previously unseen environments with very different appearance. To this end, we run the truck in three environments shown in \figLabel \ref{fig:physical_generalization}. The truck is able to consistently follow the lane in all tested environments and is responsive to commands. These and other experiments are shown in the supplementary video.

%% file: tex/conclusions.tex
\section{Discussion}

We proposed command-conditional imitation learning: an approach to learning from expert demonstrations of low-level controls and high-level commands. At training time, the commands resolve ambiguities in the perceptuomotor mapping, thus facilitating learning. At test time, the commands serve as a communication channel that can be used to direct the controller.

We applied the presented approach to vision-based driving of a physical robotic vehicle and in realistic simulations of dynamic urban environments. Our results show that the command-conditional formulation significantly improves performance in both scenarios.

While the presented results are encouraging, they also reveal that significant room for progress remains. In particular, more sophisticated and higher-capacity architectures along with larger datasets will be necessary to support autonomous urban driving on a large scale. We hope that the presented approach to making driving policies more controllable will prove useful in such deployment.

Our work has not addressed human guidance of autonomous vehicles using natural language: a mode of human-robot communication that has been explored in the literature~\cite{Broad2017,Hemachandra2015,Matuszek2014,Tellex2011,Walter2013}. We leave unstructured natural language communication with autonomous vehicles as an important direction for future work.

\section{Acknowledgements}
Antonio M. L\'opez and Felipe Codevilla acknowledge the Spanish project TIN2017-88709-R (Ministerio de Economia, Industria y Competitividad) and the Spanish DGT project SPIP2017-02237, the Generalitat de Catalunya CERCA Program and its ACCIO agency.  Felipe Codevilla was supported in part by FI grant 2017FI-B1-00162. Antonio and Felipe also thank Germ\'an Ros who proposed to investigate the benefits of introducing route commands into the end-to-end driving paradigm during his time at CVC. 